\documentclass[lettersize,journal]{IEEEtran}
\usepackage{amsmath,amsfonts}
\usepackage{algorithm}
\usepackage{array}
\usepackage{textcomp}
\usepackage{stfloats}
\usepackage{url}
\usepackage{verbatim}
\usepackage{graphicx}
\usepackage{amssymb}
\usepackage{mathtools}
\usepackage{amsthm}
\usepackage{bm}
\usepackage[american]{babel}
\usepackage[hidelinks]{hyperref}
\mathtoolsset{showonlyrefs,showmanualtags}
\usepackage{color}

\usepackage{xcolor}
\usepackage[table]{xcolor}
\usepackage{amsmath}
\usepackage{amssymb}
\usepackage{times}
\usepackage{soul}
\usepackage{url}
\usepackage[utf8]{inputenc}
\usepackage{amsthm}
\usepackage{booktabs}
\usepackage[switch]{lineno}
\usepackage{multirow}
\usepackage{bm}
\usepackage{subfig}
\usepackage{listings}
\usepackage{float}
\usepackage{xcolor}
\usepackage{booktabs}
\usepackage{graphicx}
\usepackage{multicol}
\usepackage{multirow}
\usepackage{algpseudocodex}

\definecolor{codegreen}{rgb}{0,0.6,0}
\definecolor{codegray}{rgb}{0.5,0.5,0.5}
\definecolor{codepurple}{rgb}{0.58,0,0.82}
\definecolor{backcolour}{rgb}{0.95,0.95,0.92}

\lstdefinestyle{mystyle}{
    backgroundcolor=\color{backcolour},
    commentstyle=\color{codegreen},
    keywordstyle=\color{magenta},
    numberstyle=\tiny\color{codegray},
    stringstyle=\color{codepurple},
    basicstyle=\ttfamily\footnotesize,
    breakatwhitespace=false,
    breaklines=true,
    captionpos=b,
    keepspaces=true,
    numbers=left,
    numbersep=5pt,
    showspaces=false,
    showstringspaces=false,
    showtabs=false,
    tabsize=2,
}

\definecolor{bgcolor}{rgb}{0.1, 0.1, 0.1}
\definecolor{commentcolor}{rgb}{0.5, 0.5, 0.5}
\definecolor{keywordcolor}{rgb}{0.5, 0.0, 0.5}
\definecolor{stringcolor}{rgb}{0.0, 0.5, 0.0}
\definecolor{numbercolor}{rgb}{0.0, 0.0, 0.7}
\definecolor{functioncolor}{rgb}{0.0, 0.0, 0.5}

\lstdefinestyle{github}{
    backgroundcolor=\color{bgcolor},
    basicstyle=\ttfamily\small\color{white},
    commentstyle=\color{commentcolor},
    keywordstyle=\color{keywordcolor}\bfseries,
    stringstyle=\color{stringcolor},
    numberstyle=\color{numbercolor},
    identifierstyle=\color{white},
    showstringspaces=false,
    numbers=left,
    numbersep=5pt,
    tabsize=4,
    breaklines=true,
}

\lstnewenvironment{python}[1][]
{
    
    \lstset{
    language=python,
    style=mystyle,
    #1
    }
}{}

\newcommand{\textbfu}{\underline}

\begin{document}
\bstctlcite{IEEEexample:BSTcontrol}

\clearpage
\newpage

\title{BBOPlace-Bench: Benchmarking Black-Box Optimization for Chip Placement}

\author{Ke Xue$^*$, Ruo-Tong Chen$^*$, Rong-Xi Tan$^*$, Xi Lin, Yunqi Shi, \\ Siyuan Xu, Mingxuan Yuan, Chao Qian,~\IEEEmembership{Senior Member,~IEEE}
\thanks{
The first three authors contributed equally. Chao Qian is the corresponding author (email: qianc@nju.edu.cn).

Ke Xue, Ruo-Tong Chen, Rong-Xi Tan, Xi Lin, Yunqi Shi, and Chao Qian are with the State Key Laboratory of Novel Software Technology, and also with the School of Artificial Intelligence, Nanjing University.

Siyuan Xu and Mingxuan Yuan are with Huawei Noah's Ark Lab.

This work was supported by the National Natural Science Foundation of China (624B2069), the Fundamental Research Funds for the Central Universities (14380020), the Fundamental and Interdisciplinary Disciplines Breakthrough Plan of the Ministry of Education of China (JYB2025XDXM118), and the ``111 Center'' (B26023).
}
}

\markboth{IEEE Transactions on Evolutionary Computation,~Vol.~xx, No.~x,~2026}%
{Xue \MakeLowercase{\textit{et al.}}: BBOPlace-Bench}
\maketitle

\begin{abstract}
Chip placement is a vital stage in modern chip design, and black-box optimization (BBO) has been applied to it for decades. Early BBO efforts, however, were limited by immature problem formulations and inefficient algorithm designs, leading to worse efficiency, quality, and scalability than mainstream analytical methods. Recent advances in BBO have shown strong potential, but a unified, BBO-specific benchmark for thoroughly assessing various problem formulations and BBO algorithms is lacking. To fill this gap, we propose BBOPlace-Bench, the first benchmark tailored for evaluating and developing BBO algorithms for chip placement. It integrates three BBO problem formulations and offers a modular, flexible framework that enables users to seamlessly implement, test, and compare their own algorithms. It aggregates representative modern chip cases and standardizes their formats, providing uniform and comprehensive information to support BBO optimization. Moreover, it integrates representative BBO algorithm families, including simulated annealing, population-based search (including GA, CMA-ES, and PSO), and Bayesian optimization, and systematically evaluates their performance across different problem formulations using key chip-placement metrics. We position these experiments primarily as illustrative case studies under a shared evaluation protocol, including common benchmark instances, metric definitions, evaluation pipeline, and search budgets. Under this protocol, some BBO configurations (e.g., GA under the mask-guided optimization formulation) are competitive with representative analytical and reinforcement learning baselines. BBOPlace-Bench not only facilitates the development of efficient BBO-driven solutions for chip placement but also broadens the practical application scenarios urgently needed by the BBO community.
\end{abstract}

\section{Introduction}
Chip placement is a critical yet time-consuming step in the very large-scale integrated (VLSI) design flow and significantly impacts the power, performance, and area (PPA) metrics of the final chip~\cite{mac2000industrial,markov2012progress,goldie2024chip,flowplace}.
A modern chip typically consists of numerous modules, comprising thousands of macros (i.e., individual building blocks such as memory components) and millions of standard cells (i.e., smaller fundamental elements like logic gates).
Global placement (GP) aims to place all modules, typically beginning with macros and then proceeding to standard cells. During this process, macro placement (MP) establishes a foundational layout for subsequent stages, such as standard-cell placement and routing (connecting modules using wires), thereby playing a vital role in chip design~\cite{tang2007memetic}. For example, suboptimal MP results can complicate the optimal positioning of standard cells, which may ultimately lead to unsatisfactory chip performance~\cite{vashisht2020placement}. Furthermore, inappropriate MP can cause macro blockage within the core area, inducing routing congestion, increased wirelength, and timing degradation, thereby adversely affecting overall chip performance~\cite{incre-macro,macro-regulator}.

Chip placement is inherently a black-box optimization (BBO) problem, because its objective function (measuring the PPA metrics of the final chip) has no analytical form and can only be evaluated via simulation or even physical manufacturing.
Furthermore, objective evaluation incurs high computational costs, e.g., several hours for exact simulation. To solve this expensive BBO task of chip placement, chip designers often rely on proxy metrics that reflect final chip performance to guide the optimization process~\cite{caldwell1999wirelength}, thereby significantly reducing the cost of exact evaluation. One important proxy metric is half-perimeter wirelength (HPWL), which provides an approximation of the routing wirelength and tightly correlates with routability, timing, and power, making it widely used to measure placement quality~\cite{shahookar1991vlsi,kahng2006tale,lu2015eplace}. As a result, chip placement is often formulated as the problem of minimizing HPWL subject to the constraint of zero overlap among circuit components.

There are three kinds of mainstream methods for chip placement: analytical methods, reinforcement learning (RL)-based methods, and BBO methods. Analytical placers approximate the non-differentiable HPWL using a smooth optimization objective (e.g., weighted-average wirelength~\cite{lu2015eplace,cheng2018replace}) and then optimize the approximate objective by gradient descent. Analytical placers can be accelerated by advanced AI hardware and software~\cite{lin2020dreamplace,xplace,dreamplace4}, enabling the efficient handling of the placement of millions of standard cells. Since the publication of AlphaChip in \textit{Nature}~\cite{nature-graph,alphachip}, RL has attracted growing attention as a promising approach for automated chip placement~\cite{deeppr,lai2022maskplace,lai2023chipformer,macro-regulator,fast-place}, owing to its high optimization efficiency and generalization potential. In the RL framework, chip placement is formulated as a Markov decision process, where the agent places one macro at each step and the environment transitions to the next state accordingly.

The application of BBO for chip placement has a history of several decades~\cite{murata1996vlsi,chang2000b,hong2000corner}.
Nevertheless, early attempts were hampered by immature problem formulation and inefficient algorithm design, resulting in poor placement performance. For example, the \textit{Nature} paper~\cite{nature-graph} claimed that it was ``\textit{very slow and difficult to parallelize, thereby failing to scale to the increasingly large and complex circuits of the 1990s and beyond}.'' As chip scales continue to increase, these issues become progressively more severe, preventing early BBO methods from competing with advanced analytical placers~\cite{lu2015eplace,cheng2018replace,lin2020dreamplace,xplace}.
Fortunately, recent advancements in BBO problem formulation (such as WireMask-Guided BBO~\cite{wiremask-bbo}) and efficient BBO algorithm design have highlighted the competitive potential of BBO for chip placement, with reported wirelength reductions of 30\% and 23\% compared to the analytical placer DREAMPlace~\cite{lin2020dreamplace} and RL placer MaskPlace~\cite{lai2022maskplace}, respectively.
Despite these advancements, the field lacks a unified benchmark for thoroughly assessing various BBO problem formulations and BBO algorithms, which hinders systematic comparisons and progress, underscoring a critical need for a standardized platform.

\begin{figure*}[htbp]\centering
\includegraphics[width=0.99\linewidth]{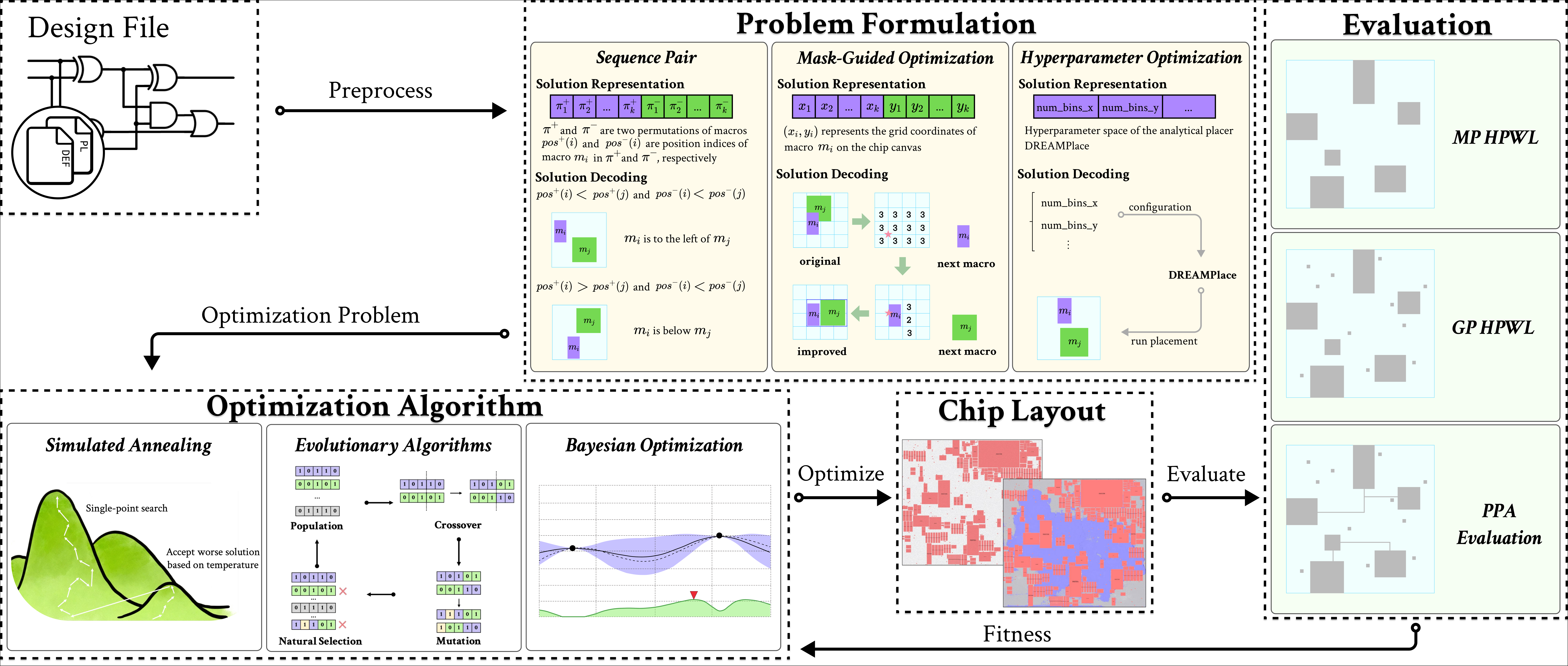}
\caption{Illustration of BBOPlace-Bench. It decouples three core components, i.e., problem formulation, optimization algorithm, and evaluation, making it easier to test different BBO algorithms under customized settings. The three problem formulations are: 1) Sequence Pair: using two permutations to capture positional relationships of macros. 2) Mask-Guided Optimization: representing macros by grid coordinates and using wire-mask-guided decoding to adjust macro positions to grids with minimal incremental HPWL. 3) Hyperparameter Optimization: optimizing the advanced analytical placer DREAMPlace's hyperparameters (e.g., learning rate and target density) over mixed (discrete and continuous) search spaces, where each configured DREAMPlace run serves as solution decoding and generates a macro placement. Evaluation metrics include: MP-HPWL (a fast proxy metric for macro wirelength), GP-HPWL (an accurate but costly metric for wirelength of macros and standard cells),
and PPA (industrial chip metrics obtained via EarlyGlobalRoute).}
\label{fig:overview}
\end{figure*}

To fill this gap, we propose BBOPlace-Bench, the first benchmark designed specifically for BBO users to tackle chip placement tasks. Our motivation for focusing on BBO is primarily problem-driven: chip placement naturally fits the black-box setting, and it is also an important real-world application that helps address an urgent need in the BBO community for practical benchmarks. As illustrated in Figure~\ref{fig:overview}, BBOPlace-Bench establishes a modular, decoupled framework that separates three core components—problem formulation, optimization, and evaluation—enabling users to easily test diverse BBO algorithms under customized settings.
The framework supports three distinct problem formulations, each with a tailored process that maps a genotype solution to module positions: 1) Sequence pair (SP) utilizes pairs of permutation sequences to represent the relative relationships of modules, and employs the longest common subsequence algorithm for solution decoding;
2) Mask-guided optimization (MGO) uses grid coordinates for module representation and employs a mask-guided mechanism for solution decoding;
3) Hyperparameter optimization (HPO) optimizes the hyperparameters of the advanced analytical placer DREAMPlace~\cite{lin2020dreamplace, dreamplace4} over mixed (discrete and continuous) search spaces, where each configured DREAMPlace run serves as solution decoding and generates a module placement.
BBOPlace-Bench integrates representative BBO algorithm families, including simulated annealing (SA)~\cite{sa,murata1996vlsi}, population-based search~\cite{back1996evolutionary,elbook,hansen2015evolution,cmaes,pso,gong2015genetic} (including genetic algorithms, evolution strategies via CMA-ES, and particle swarm optimization (PSO)), and Bayesian optimization (BO)~\cite{bosurvey1,bobook}, which interact with the problem formulation to optimize layouts using fitness feedback.
For evaluation, BBOPlace-Bench standardizes two industrial chip suites (ISPD 2005~\cite{nam2005ispd2005} and ICCAD 2015~\cite{iccad15}) and offers multiple metrics: MP-HPWL (i.e., HPWL of macros) for general BBO scenarios and GP-HPWL (i.e., HPWL of macros and all standard cells) for expensive evaluation settings, alongside PPA evaluation through EarlyGlobalRoute. This architecture systematically decouples components to accelerate research into how problem formulation, algorithm design, and evaluation metrics interact in advancing BBO for chip placement.

To demonstrate the usage and versatility of BBOPlace-Bench, we conduct comprehensive experiments on various chip cases, systematically evaluating five representative BBO algorithms (SA, GA, CMA-ES, PSO, BO) across the three aforementioned problem formulations.
Our experimental design targets two core evaluation goals: 1) minimizing MP-HPWL for general BBO evaluation scenarios and optimizing GP-HPWL to assess performance under expensive evaluation scenarios; 2) benchmarking against representative analytical methods (e.g., DREAMPlace~\cite{lin2020dreamplace,dreamplace4}) and RL approaches (e.g., AlphaChip~\cite{nature-graph,alphachip}, MaskPlace~\cite{lai2022maskplace}, and EfficientPlace~\cite{fast-place}) under a shared reporting protocol. These experiments are primarily illustrative case studies that demonstrate how BBOPlace-Bench can be used; they are not intended as an exhaustive ranking across all placement workflows. Under the reported metrics, budgets, and seeds, the results still reveal useful contrasts among formulations and methods and highlight where BBO can be competitive on proxy objectives, while also exposing limitations such as high-dimensional BO challenges.

BBOPlace-Bench not only provides a practical benchmarking infrastructure for chip placement and facilitates the development of efficient BBO-driven solutions in this field, but also expands the application scenarios of BBO algorithms and broadens their practical use in chip design.
Our contributions can be summarized as follows:
\begin{itemize}
\item We propose the first BBO benchmark BBOPlace-Bench for chip placement, providing an important, real-world, and challenging task for BBO algorithms. BBOPlace-Bench integrates three problem formulations (SP, MGO, and HPO) of BBO for chip placement and five representative BBO algorithms (SA, GA, CMA-ES, PSO, and BO), and supports evaluation with three kinds of metrics (MP-HPWL, GP-HPWL, and PPA) on two standard industrial chip suites (ISPD 2005 and ICCAD 2015).
\item We decouple problem formulation, optimization algorithms, and evaluation, making BBOPlace-Bench convenient for researchers in the BBO community to compare their own algorithms under a clear protocol.
\item We conduct extensive empirical studies, including MP-HPWL and GP-HPWL evaluation on ISPD 2005 and ICCAD 2015, as well as PPA evaluation on ICCAD 2015, and provide detailed discussions on different problem formulations and BBO algorithms. Under common benchmark instances, metric definitions, evaluation pipeline, and search budgets, MGO and HPO generally obtain higher average ranks than SP.
\item This flexible framework not only facilitates work on the critical chip placement problem but also provides an important real-world benchmark urgently needed by the BBO community. By covering challenging scenarios such as high-dimensional and expensive optimization, BBOPlace-Bench can promote BBO research and expand the field's practical impact.
\end{itemize}

\section{Background}

\subsection{Black-Box Optimization}\label{sec2.2}
Black-box optimization (BBO) refers to the process of optimizing an objective function whose mathematical formulation and internal structure are unknown or inaccessible, relying solely on input-output evaluations to iteratively search for optimal solutions~\cite{dfo-book1,bobook,dfo-book2}.
BBO algorithms can be organized in different ways, and no single taxonomy is universally agreed upon. In this paper, we adopt a pragmatic exposition scaffold by search style: single-point iterative search (SA), population-based search (represented by GA, CMA-ES, and PSO), and model-based sequential search (BO).
SA~\cite{sa,murata1996vlsi} is a stochastic optimization algorithm that performs search by iteratively mutating the candidate solution. A key characteristic of SA is its ability to accept worse solutions with a probability that is dynamically controlled by a predefined cooling schedule (referred to as temperature), which enables the algorithm to escape from local optima and gradually approach high-quality solutions. However, traditional SA has inherent drawbacks, such as a slow convergence rate, which primarily stems from its single-point search mechanism that explores the solution space sequentially.
Population-based methods maintain a set of candidate solutions (i.e., a population or swarm) and update them in parallel, thereby enhancing exploration efficiency and mitigating limitations of single-point search strategies. In this paper, we use GA, CMA-ES~\cite{hansen2015evolution,cmaes}, and PSO~\cite{pso,gong2015genetic} as representative built-in methods.
BO~\cite{bosurvey1,bobook} is a widely used sample-efficient method for expensive BBO problems. At each iteration, BO fits a surrogate model, typically a Gaussian process~\cite{gpml}, to approximate the objective function, and maximizes an acquisition function (e.g., probability of improvement~\cite{Krushner64PI}, expected improvement~\cite{ei2}, or upper confidence bound~\cite{gpucb}) to determine the next query point, automatically balancing exploration and exploitation.

Although BBO has many practical applications, \textit{real-world} tasks addressed in academic research remain limited, primarily focusing on problems such as hyperparameter optimization of machine learning algorithms~\cite{HPOB,nas101} and robotic control~\cite{mujoco}. This paper formulates the important and challenging problem of chip placement within a BBO-friendly framework, thereby expanding the application scope of BBO. The framework is user-friendly and facilitates the integration of various advanced BBO algorithms.

\subsection{Chip Placement}\label{sec2.1-placement}

Chip placement is a vital stage in modern chip design.
The main input of chip placement is a netlist $\mathcal{N}=(V,E)$, where $V$ denotes the set of modules in the chip, including macros and standard cells, together with their basic geometric information such as width and height, and $E$ denotes the set of nets, where each net connects a group of modules and represents their interconnection relationship.
Given a netlist, a fixed canvas layout, and a library of modules, a chip placement method is expected to determine the appropriate physical locations of movable modules on the canvas such that an objective used to measure chip performance (e.g., total wirelength) can be optimized.
Given $k$ modules to be placed $\{m_i\}_{i=1}^k$, a legal chip placement solution $\bm{s}=\{(x_i, y_i)\}_{i=1}^k\in \mathbb R^{2k}$ includes the positions of all the modules, where $x_i$ and $y_i$ denote the horizontal and vertical physical coordinates of module $m_i$, respectively.
As discussed above, a typical formulation of chip placement is to minimize the total HPWL of all the nets while satisfying the cell density constraint, which is formulated as follows:
\begin{align}\label{eq:obj}
\min_{\bm{s}} \mathrm{HPWL}(\bm{s}) = &\min_{\bm{s}} \sum_{e\in E}\mathrm{HPWL}_e(\bm{s}), \\
& \text{s.t.} \ D(\bm{s}) \leq \epsilon,
\end{align}
where $D$ denotes the density (such as rectangular uniform wire density~\cite{rudy}), and $\epsilon$ is a threshold of density.

\begin{figure}[t!]\centering
\includegraphics[width=0.6\linewidth]{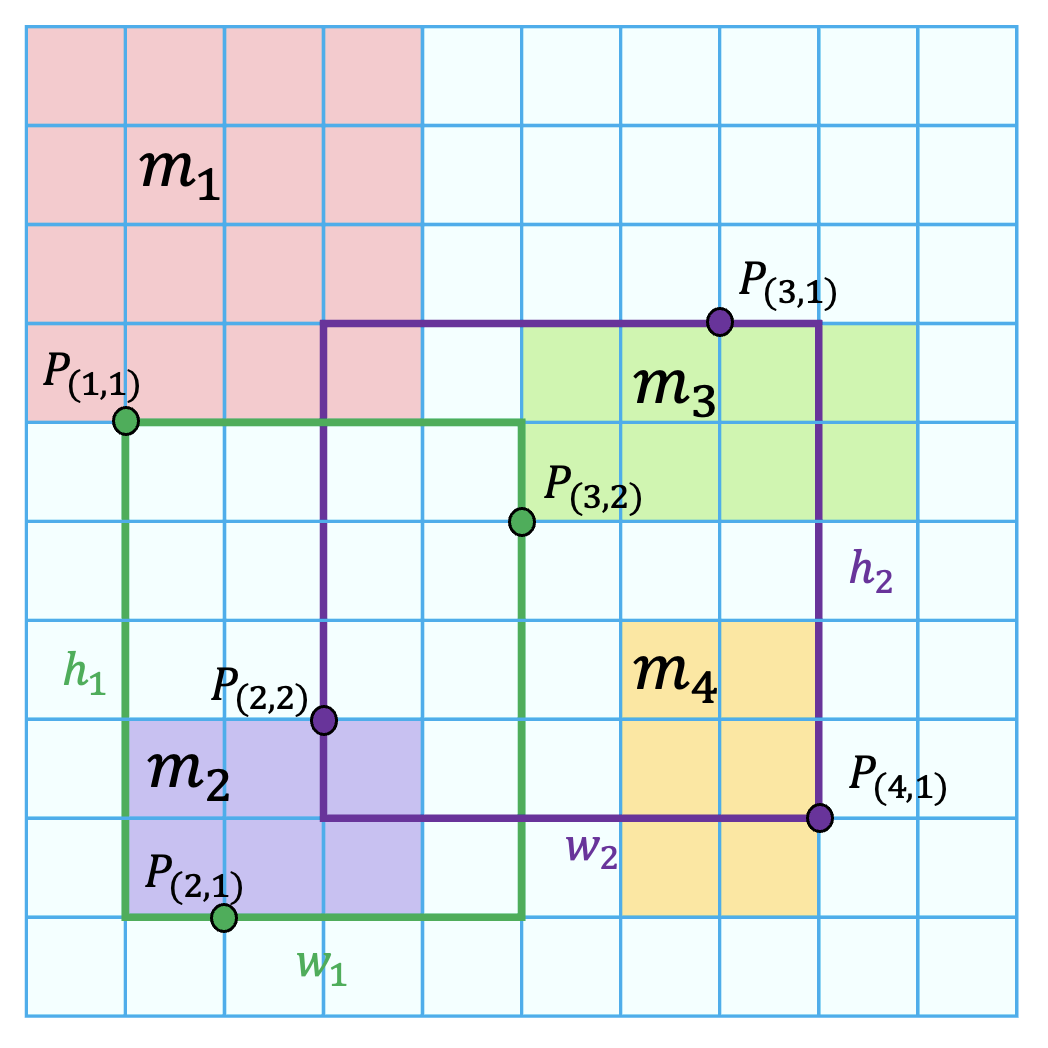}
\caption{Example illustration of calculating HPWL.
}\label{fig:hpwl}
\end{figure}

$\mathrm{HPWL}_e$ in Eq.~\eqref{eq:obj} is the HPWL of net $e$, which is defined as:
\begin{align}\label{eq:hpwle}
\mathrm{HPWL}_e(\bm{s}) = &\,(\max\nolimits_{P\in e} P_x - \min\nolimits_{P\in e} P_x) \\
&+ (\max\nolimits_{P\in e} P_y - \min\nolimits_{P\in e} P_y),
\end{align}
where $P \in e$ denotes a pin of a module in net $e$. Pins serve as input and output interfaces for modules and should be connected by wires; $P_x$ and $P_y$ denote the horizontal and vertical coordinates of pin $P$, respectively. Intuitively, $\mathrm{HPWL}_e$ is calculated as the half perimeter of the rectangle that encloses all the pins in net $e$. To help BBO researchers understand HPWL, we provide an example illustration of a 2D chip canvas in Figure~\ref{fig:hpwl}. There are four modules $\{m_i\}_{i=1}^4$ on the chip canvas, where $P_{(i,j)}$ denotes the $j$-th pin of module $m_i$. Net $e_1$ (colored in green) connects $m_1$, $m_2$, and $m_3$, while net $e_2$ (colored in purple) connects $m_2$, $m_3$, and $m_4$; the two colored bounding boxes correspond to these two nets, and the total HPWL is obtained by summing their widths and heights based on Eqs.~\eqref{eq:obj} and~\eqref{eq:hpwle}, i.e., $w_1+h_1+w_2+h_2=4+5+5+5=19$.

The calculation of HPWL is non-differentiable and the optimization of Eq.~\eqref{eq:obj} is NP-complete~\cite{lu2015eplace}. Solving such a problem is extremely challenging. There are three mainstream methods for chip placement, i.e., analytical methods, reinforcement learning-based methods, and black-box optimization methods, which will be introduced in the next section.

\subsection{Existing Methods and Benchmarks}

\paragraph{Analytical Methods}
Analytical methods reformulate the optimization objective in Eq.~\eqref{eq:obj} as a smooth approximation because the original HPWL calculation is non-differentiable. They typically use the weighted-average wirelength model~\cite{hsu2011tsv} as a smooth surrogate to approximate the gradient, which allows the placement problem to be solved efficiently using gradient descent. Specifically, the smoothed objective is formulated as:
\begin{align}\label{eq:smoothed-hpwl}
\min_{\bm{s}} \sum_{e\in E} \mathrm{WA}_e(\bm{s}) + \lambda \cdot D(\bm{s}),
\end{align}
where $\mathrm{WA}_e(\bm{s}) = \mathrm{WA}_e(\bm{x}) + \mathrm{WA}_e(\bm{y})$, and $\mathrm{WA}_e(\bm{x})$ is defined as:
\begin{equation}
\mathrm{WA}_e(\bm{x}) = \frac{\sum_{P\in e} P_x e^{P_x/\gamma}}{\sum_{P\in e} e^{P_x/\gamma}} - \frac{\sum_{P\in e} P_x e^{-P_x/\gamma}}{\sum_{P\in e} e^{-P_x/\gamma}},
\end{equation}
and $\mathrm{WA}_e(\bm{y})$ is defined similarly. The parameter $\gamma$ controls the approximation accuracy, and $\lambda$ is the density penalty coefficient.
They typically place macros and standard cells simultaneously, and can be roughly categorized into the following two kinds of approaches~\cite{essential-issues-in-analytical}.
Quadratic placement~\cite{ripple,polar} iterates between an unconstrained quadratic programming phase to minimize wirelength and a heuristic spreading phase to remove overlaps.
Nonlinear placement~\cite{chen2008ntuplace3,lu2015eplace,cheng2018replace} formulates a nonlinear optimization problem and tries to directly solve it with gradient descent methods, which can achieve better solution quality but usually at a higher computational cost.
Recently, there has been extensive attention on GPU-accelerated nonlinear placement methods.
For example, DREAMPlace~\cite{lin2020dreamplace,dreamplace4} transforms the nonlinear placement problem of optimizing Eq.~\eqref{eq:obj} into a neural network training problem and solves it by gradient descent with GPU acceleration, enabling ultra-high parallelism and producing state-of-the-art analytical placement quality.
However, the existing analytical approaches still suffer from the following two drawbacks:
1) Analytical placers optimize complex objectives by approximating gradients, a process prone to getting stuck in local optima and even encountering failure cases~\cite{lai2022maskplace,xue2024escaping};
2) Advanced analytical placers involve numerous hyperparameters that require tuning, and optimizing them effectively remains non-trivial~\cite{agnesina2023autodmp,dacdmp}.

\paragraph{Reinforcement Learning-based Methods}
RL has become a popular approach for chip placement since AlphaChip~\cite{nature-graph,alphachip} was published in \textit{Nature} in 2021 and demonstrated RL-based chip placement. AlphaChip uses a two-stage pipeline: an RL agent places the macros one by one in the first stage, and a placement tool handles the remaining standard cells in the second stage.
For the first stage, AlphaChip divides the chip canvas into discrete grids, then orders macros by physical size (typically prioritizing larger macros first), and finally learns to place one macro per time step until all are placed. The detailed Markov decision process (MDP) is formulated as follows:
States encode key placement information, such as netlist metadata; actions are valid grid coordinates for placing the current macro without violating constraints; rewards are zero for all intermediate actions and a negatively weighted sum of proxy wirelength, congestion, and density for the final action that produces a complete placement.
Since then, many studies on RL for chip placement have sought to address AlphaChip's limitations, such as enhancing state-representation capabilities~\cite{deeppr,cheng2022the}, constructing dense rewards~\cite{lai2022maskplace,macro-regulator}, incorporating offline training~\cite{lai2023chipformer}, and leveraging efficient planning~\cite{fast-place}, with recent studies showing competitive or better performance compared with analytical placers. However, RL-based placement methods still face challenges in generalizing to different circuit designs and require substantial computational resources for training, which may limit their practical applicability in industrial settings.

\paragraph{Black-box Optimization Methods}
BBO has been used for chip placement for three decades. Earlier approaches, such as the sequence pair (SP~\cite{murata1996vlsi,bomp}), suffer from poor scalability due to an inefficient rectangular packing formulation, which requires compact placements that do not conform to advanced chip-design standards and involves an enormous search space as well as complex transformations from BBO solutions to legal placement results. Recently, some BBO methods have made significant progress by changing the search space.
AutoDMP~\cite{agnesina2023autodmp} improves DREAMPlace by using Bayesian optimization to explore the hyperparameter configuration space and shows remarkable performance on multiple benchmarks.
WireMask-BBO~\cite{wiremask-bbo} adopts a wire-mask-guided greedy mapping from genotype solutions to phenotype solutions, and can be equipped with any BBO algorithm, demonstrating superior performance over packing-based, RL, and analytical methods. In this paper, our proposed BBOPlace-Bench integrates these BBO problem-formulation approaches into a unified benchmark for easier comparison and development of BBO algorithms for chip placement.

\paragraph{Benchmarks}
Recently, several benchmarks have been proposed for AI in EDA.
Competitions in the EDA field, such as the ISPD~\cite{nam2005ispd2005} and ICCAD~\cite{iccad15} contests, provide datasets with basic chip information, including fundamental netlist information for several chips. However, they do not offer placement methods, and researchers need to extract the data themselves and implement their own methods. CircuitNet~\cite{circuitnet1,circuitnet2} focuses on providing multi-modal data for chip performance prediction tasks, enhancing predictive capability through diverse data modalities. ChiPBench~\cite{chipbench} emphasizes the entire end-to-end EDA workflow, providing complete files for each case and necessary design kits, thereby offering a comprehensive dataset that supports all stages of design.
In contrast, our proposed BBOPlace-Bench aims to provide a unified and user-friendly benchmark that is specifically designed for BBO in chip placement research, allowing researchers in the BBO community to focus solely on BBO algorithm design and encouraging the expansion of BBO applications in the emerging and critical field of chip placement.

\section{BBOPlace-Bench}\label{sec:3}

In this section, we introduce our proposed BBO benchmark, BBOPlace-Bench, for chip placement.
Figure~\ref{fig:overview} gives an illustration of our BBOPlace-Bench. Without BBOPlace-Bench, applying BBO to chip placement involves significant engineering barriers, including parsing complex industrial dataset formats (e.g., Bookshelf and LEF/DEF), designing legal mappings from BBO solutions to physical layouts, and integrating complex EDA tools for evaluation. BBOPlace-Bench addresses these obstacles by unifying the preprocessing, decoding, and evaluation pipelines, allowing BBO researchers to focus solely on algorithm design.
We first introduce how to bridge existing chip placement cases, i.e., ISPD 2005~\cite{nam2005ispd2005} and ICCAD 2015~\cite{iccad15}, with BBO in Section~\ref{sec:bench-chips}. Then, we introduce the three major components in our benchmark, i.e., problem formulation, optimization algorithm, and evaluation in Sections~\ref{sec:3-2}, \ref{sec:3-3}, and \ref{sec:3-4}, respectively. We finally introduce the BBO user-friendly interfaces and give a code example in Section~\ref{sec:3-5}.

\subsection{Preprocessing: Bridging Chip Placement and BBO}\label{sec:bench-chips}
With the rapid development of EDA, datasets for chip design have undergone significant changes in structure and format. Early datasets, such as ISPD 2005~\cite{nam2005ispd2005}, use a simplified \textit{Bookshelf} format. Compared with later industrial formats such as \textit{LEF/DEF}, this simplified format lacks important design information (e.g., detailed floorplan descriptions and routing-related constraints) needed for realistic downstream analysis and evaluation. In contrast, later datasets such as ICCAD 2015~\cite{iccad15} are much more complex, offering \textit{LEF/DEF} format along with other necessary files. To address the challenges posed by different dataset formats, we provide interfaces that are compatible with both \textit{Bookshelf} and \textit{LEF/DEF} formats and capable of processing them. From these processed data, we extract the essential information required for the placement stage, making it easier to implement problem-formulation and evaluation components for BBO.

\subsection{Problem Formulations}\label{sec:3-2}
This section presents three problem-formulation approaches for BBO in chip placement, i.e., how to map a BBO solution (genotype) to a valid chip placement solution (phenotype).

\subsubsection{Sequence Pair (SP)}
SP is a traditional combinatorial way of encoding solutions in chip placement~\cite{murata1996vlsi}, which represents a chip placement solution by a pair of permutations $\bm{s} = (\pi^{+}, \pi^{-})$ of $k$ modules \(\{m_{i}\}_{i=1}^{k}\). SP extracts relative relationships between modules from the permutations and then determines their coordinates on the chip canvas by the longest common subsequence (LCS) algorithm. The detailed SP problem-formulation flow is shown in Algorithm~\ref{alg:sequence_pair_formulation}.
\begin{algorithm}[t!]
    \caption{Sequence pair formulation for chip placement}
    \textbf{Input} Netlist $\mathcal{N}=(V,E)$, set of modules $\{m_i\}_{i=1}^k$ to be placed, SP solution $\bm{s}=(\pi^+,\pi^-)$ \\
    \textbf{Output} Modules' coordinates $\{(x_i, y_i)\}_{i=1}^k$
    \label{alg:sequence_pair_formulation}
    \begin{algorithmic}[1]
        \State Initialize empty horizontal order array $H$ and vertical order array $V$
        \State Compute position indices $pos^+(i)$ in $\pi^+$ and $pos^-(i)$ in $\pi^-$ for each module $m_i$
        \For{each module pair $(m_i, m_j)$}
            \If{$pos^+(i) < pos^+(j)$ \textbf{and} $pos^-(i) < pos^-(j)$}
                \State $m_i$ is to the left of $m_j$, add $(m_i, m_j)$ to $H$
            \ElsIf{$pos^+(i) > pos^+(j)$ \textbf{and} $pos^-(i) < pos^-(j)$}
                \State $m_i$ is below $m_j$, add $(m_i, m_j)$ to $V$
            \EndIf
        \EndFor
        \State Determine horizontal and vertical orders using longest common subsequence on $H$ and $V$
        \State Calculate $x$-coordinates based on horizontal order and $y$-coordinates based on vertical order
        \State Adjust coordinates to ensure no overlaps and minimal area
    \end{algorithmic}
    \textbf{Return} Modules' coordinates $\{(x_i, y_i)\}_{i=1}^k$
\end{algorithm}

For any two modules \(m_{i}\) and \(m_{j}\), their relative positions (i.e., left, right, down, up) are determined by comparing their indices in \(\pi^{+}\) and \(\pi^{-}\). Specifically, for each module \(m_{i}\), we compute its position indices \(pos^{+}(i)\) in \(\pi^{+}\) and \(pos^{-}(i)\) in \(\pi^{-}\).
The relative relationships are established as follows:
If \(pos^{+}(i) < pos^{+}(j)\) and \(pos^{-}(i) < pos^{-}(j)\), \(m_{i}\) is to the left of \(m_{j}\);
If \(pos^{+}(i) > pos^{+}(j)\) and \(pos^{-}(i) < pos^{-}(j)\), \(m_{i}\) is below \(m_{j}\).
These relationships are stored in the horizontal order array \(H\) and vertical order array \(V\). Then, the LCS algorithm is applied to \(H\) and \(V\) to determine the horizontal and vertical orders, from which \(x\) and \(y\) coordinates of the modules are calculated. Finally, these coordinates are adjusted to avoid overlaps and minimize area, yielding the final placement solution \( \{(x_{i}, y_{i})\}_{i=1}^{k}\), where $(x_{i}, y_{i})$ is the physical coordinate of module $m_i$ on the chip canvas. SP provides a structured combinatorial framework for encoding module placements, leveraging permutation-based relationships to ensure minimal area placement with no further adjustments needed. Its formulation is intuitive for modeling relative positions, making it suitable for problems where positional constraints dominate. However, the permutation search space grows exponentially with the number of modules. Additionally, extracting the relative relationships for all module pairs requires $O(k^2)$ time complexity due to the double-loop structure, leading to high computational costs and poor scalability for large designs~\cite{nature-graph}.

\subsubsection{Mask-Guided Optimization (MGO)} Since the chip canvas is two-dimensional, it is natural to treat it as a grid and represent solutions using coordinates. Under this grid representation, a chip placement solution $\bm{s}$ is defined by the coordinates of all modules $\{m_i\}_{i=1}^k$, i.e., $\bm{s}=\{(x_i,y_i)\}_{i=1}^k$, where $(x_i, y_i)$ denotes the coordinates of module $m_i$ on the chip canvas. However, direct optimization in this search space makes it difficult to efficiently find a solution that has a small HPWL value and satisfies the non-overlapping constraint~\cite{net-based-force,lai2022maskplace}. To improve the optimization efficiency and ensure constraint satisfaction, Shi et al.~\cite{wiremask-bbo} proposed a wire-mask-guided greedy procedure, which converts \textit{any given solution} into a valid, high-quality placement result. The detailed algorithmic flow is provided in Algorithm~\ref{alg:wiremask_guided}. For each module $m_i$, we first compute the total area of all modules connected to it through nets; modules are then permuted in descending order based on these areas to determine the adjustment priority, as modules with larger connected areas are more critical to placement quality. When placing each module in this order, a wire mask~\cite{lai2022maskplace} is generated, which records the incremental HPWL caused by placing the current module on each candidate grid. Grids that would cause overlap or exceed the canvas boundary are excluded, enabling the MGO formulation to efficiently satisfy the non-overlapping constraint. The module is then moved to the grid that minimizes the incremental HPWL; in case of ties, the grid closest to the original position of the module (as recorded in the MGO solution) is selected. This sequential, greedy adjustment ensures both small HPWL and non-overlapping placement, balancing quality and legality effectively.
\begin{algorithm}[t!]
    \caption{Mask-guided optimization formulation for chip placement}
    \textbf{Input} 
    Netlist $\mathcal{N}=(V,E)$, 
    set of modules $\{m_i\}_{i=1}^k$ to be placed,
    MGO solution \( \bm{s}=\{(x_i, y_i)\}_{i=1}^k \) \\
    \textbf{Output} Modules' coordinates $\{(x'_i, y'_i)\}_{i=1}^k$
    \label{alg:wiremask_guided}
    \begin{algorithmic}[1]
        \State Initialize chip canvas as \( n \times n \) grids
        \State Order macros $\{m_i\}_{i=1}^k$ in descending order of their areas of all the connected modules
        \For{each macro \( m_i \) in the ordered list}
            \State Generate wire mask \( W_i \) that records incremental HPWL by placing \( m_i \) in each grid
            \State Exclude grids causing overlap or exceeding canvas boundaries from \( W_i \)
            \State Find the best grids \( W^*_i \) with minimum incremental HPWL 
            \State Select the grid $(x'_i, y'_i)$ within \( W^*_i \) that is nearest to \( m_i \)'s original position $(x_i,y_i)$
        \EndFor
    \end{algorithmic}
    \textbf{Return} Modules' coordinates $\{(x'_i, y'_i)\}_{i=1}^k$
\end{algorithm}

\subsubsection{Hyperparameter Optimization (HPO)}
Although analytical placers are currently the mainstream methods for chip placement, they usually involve numerous hyperparameters that must be tuned carefully, and inappropriate adjustments can significantly degrade performance~\cite{agnesina2023autodmp}. BBO algorithms have proven effective for HPO and have achieved excellent performance on various tasks. Therefore, an effective problem formulation of BBO for chip placement~\cite{agnesina2023autodmp} is to use BBO to tune the hyperparameters of the state-of-the-art analytical placer, e.g., DREAMPlace~\cite{lin2020dreamplace,dreamplace4}. In the HPO formulation, we set the search space for chip placement as the hyperparameter space $\Theta$ of DREAMPlace, with details shown in Table~\ref{tab:hpo-search-space}. The first part in Table~\ref{tab:hpo-search-space} consists of the general placement configurations, and the second part includes the configurations at each iteration of DREAMPlace. The mixed search space requires specialized operators for handling both discrete and continuous variables. The detailed HPO formulation is shown in Algorithm~\ref{alg:hpo_algorithm}, where running a configured DREAMPlace serves as solution decoding and generates a module placement.

Note that the three BBO formulations in BBOPlace-Bench involve different search spaces: permutation, continuous, and mixed (both continuous and discrete) search spaces, which support the development of various types of BBO algorithms. The search space dimensionality is directly determined by the formulation and the number of macros $k$: SP requires two permutations of length $k$, MGO has a dimensionality of $2k$ (the $x$ and $y$ coordinates for each macro), while HPO has a fixed dimensionality of 15 (the number of hyperparameters tuned, as shown in Table~\ref{tab:hpo-search-space}). As the number of macros can be specified arbitrarily in our BBOPlace-Bench, the SP and MGO formulations can also be used for research on high-dimensional BBO, which is a popular recent topic in BBO.

\begin{table}[t!]
\caption{Search space of HPO in BBOPlace-Bench.}
\label{tab:hpo-search-space}
\centering
\resizebox{\linewidth}{!}{
\begin{tabular}{c|c|c}
\toprule
HPO search space                        & Type       & Range                                  \\ \midrule

GP\_num\_bins\_x                        & discrete   & {\{}1024, 2048{\}}                       \\
GP\_num\_bins\_y                        & discrete   & {\{}1024, 2048{\}}                       \\
GP\_optimizer                           & discrete   & {\{}``adam", ``nesterov"{\}}               \\
GP\_wirelength                          & discrete   & {\{}``weighted\_average", ``logsumexp"{\}} \\
GP\_learning\_rate                      & continuous & {[}0.001, 0.01{]}                      \\
GP\_Llambda\_density\_weight\_iteration & continuous & {[}1, 3{]}                             \\
GP\_Lsub\_iteration                     & continuous & {[}1, 3{]}                             \\
GP\_learning\_rate\_decay               & continuous & {[}0.99, 1.0{]}                        \\
stop\_overflow                          & continuous & {[}0.06, 0.1{]}                        \\
target\_density                         & continuous & {[}0.8, 1.2{]} \\ \midrule
RePlAce\_LOWER\_PCOF                    & continuous & {[}0.9, 0.99{]}                        \\
RePlAce\_UPPER\_PCOF                    & continuous & {[}1.02, 1.15{]}                       \\
RePlAce\_ref\_hpwl                      & continuous & {[}150000, 550000{]}                   \\
density\_weight                         & continuous & {[}1e-6, 1e-4{]}                       \\
gamma                                   & continuous & {[}1, 4{]}                             \\
\bottomrule                        
\end{tabular}}
\end{table}

\begin{algorithm}[t!]
    \caption{Hyperparameter optimization formulation for chip placement}
    \textbf{Input} 
    Netlist $\mathcal{N}=(V,E)$, 
    set of modules $\{m_i\}_{i=1}^k$ to be placed,
    analytical placer $\mathcal P$,
    HPO solution $\bm \theta=(\theta_1,\dots,\theta_N) \in \Theta$ \\
    \textbf{Output} Modules' coordinates $\{(x_i, y_i)\}_{i=1}^k$
    \label{alg:hpo_algorithm}
    \begin{algorithmic}[1]
        \State Set analytical placer (e.g., DREAMPlace) $\mathcal P_{\bm \theta}$ with the HPO solution $\bm \theta=(\theta_1,\dots,\theta_N)$
        \State Run analytical placement process: $\bm s \leftarrow \mathcal P_{\bm \theta}(\mathcal{N}, \{m_i\}_{i=1}^k)$
    \end{algorithmic}
    \textbf{Return} Modules' coordinates $\bm s =\{(x_i, y_i)\}_{i=1}^k$
\end{algorithm}

\subsection{Optimization Algorithms}\label{sec:3-3}

The literature proposes many ways to categorize black-box optimization (BBO) algorithms, and no single taxonomy is universally agreed upon. For clarity, this paper adopts a simple exposition scaffold organized by how search proceeds: \emph{single-point} iterative search, illustrated by simulated annealing (SA); \emph{population-based} search, which maintains and updates multiple candidates in parallel and is instantiated here by genetic algorithms (GA), CMA-ES, and particle swarm optimization (PSO) as representative built-in methods; and \emph{model-based} sequential search via a surrogate and an acquisition function, illustrated by Bayesian optimization (BO). These families are widely used and highlight complementary mechanisms, but the grouping is \emph{not} exhaustive; additional optimizers can be integrated through the provided interfaces. Default built-in algorithms are summarized as follows:
\begin{itemize}
    \item SA~\cite{sa} is a classic approach in chip placement~\cite{murata1996vlsi}. By mimicking the cooling process of metals, it effectively explores the search space, balancing exploration and exploitation to minimize an objective function. Its ability to escape from local minima makes it particularly valuable in optimizing complex layouts.
    \item Population-based search methods~\cite{back1996evolutionary}. In this paper, we implement three representative population-based algorithms: GA, CMA-ES, and PSO.
    \begin{itemize}
        \item GA is a basic population-based search method, which updates a set of solutions (i.e., population) through iterative crossover, mutation, and selection. We implement various operators to handle different types of search spaces. Details can be found in Section~\ref{sec:exp-settings}.
        \item CMA-ES is a representative method used in the field of continuous BBO. We integrate pycma\footnote{\url{https://github.com/CMA-ES/pycma}}, a popular implementation of CMA-ES~\cite{cmaes} in Python, into our benchmark. It not only provides a basic implementation of CMA-ES but also includes numerous advanced features suitable for high-dimensional optimization and many other scenarios.
        \item PSO~\cite{pso} is a population-based optimization technique inspired by the social behavior of bird flocks or fish schools. It maintains a swarm of particles that explore the search space by following their own best known position as well as the swarm's best known position. PSO is particularly effective for continuous optimization problems and has been successfully applied to various engineering problems.
    \end{itemize}
    \item BO~\cite{bobook} is a model-based sequential search via a surrogate and an acquisition function, which is sample-efficient for expensive BBO problems. We integrate one of the most popular BO frameworks, BoTorch~\cite{botorch}\footnote{\url{https://github.com/pytorch/botorch}}, into our benchmark. BoTorch leverages GPUs for efficient Gaussian process fitting and inference, and includes a wide range of advanced BO algorithms, e.g., TurBO~\cite{turbo} and SAASBO~\cite{sassbo}.
\end{itemize}

\subsection{Evaluation}\label{sec:3-4}
As an important part of the EDA process, chip placement has many evaluation metrics and procedures. In our benchmark, we employ the following three methods.

\paragraph{Macro Placement HPWL}
The chip placement problem can be divided into two successive stages~\cite{agnesina2023autodmp}: Macro placement (MP) and standard cell placement. That is, macros are first placed, and then the positions of standard cells are determined. MP heavily influences the subsequent placement of standard cells, and poor MP might make it challenging to place standard cells optimally, leading to unsatisfactory chip performance. Thus, MP-HPWL, i.e., the HPWL value over all macros, is an important metric for evaluating the quality of chip placement.

\paragraph{Global Placement HPWL}
GP-HPWL is the HPWL value of all the macros and standard cells. Compared to MP-HPWL, GP-HPWL is more closely related to the final chip performance but computing it requires placing standard cells and therefore takes much more time. In BBOPlace-Bench, after obtaining the positions of macros through different problem formulations and optimization algorithms, if GP-HPWL evaluation is required, we fix the already placed macros and then place the standard cells with DREAMPlace~\cite{lin2020dreamplace} to obtain GP-HPWL. The numerical value of GP-HPWL is typically one or two orders of magnitude larger in metric scale than MP-HPWL because GP-HPWL includes both macros and standard cells. As the number of macros is much smaller than the number of standard cells (several hundreds versus millions), MP-HPWL is cheaper to evaluate and more suitable as a surrogate metric for BBO algorithms. In our BBOPlace-Bench framework, the GP-HPWL interface can be called independently, in which case the problem can be treated as an expensive BBO problem. Detailed empirical wall-clock time comparison is provided in the supplementary material.

\paragraph{PPA Evaluation}
The entire chip-design process is lengthy and complex. Proxy metrics, such as MP-HPWL and GP-HPWL, may not accurately reflect the final chip performance, namely the power, performance, and area (PPA) metrics.
In this paper, once the global placement results (i.e., the positions of all macros and standard cells) are obtained, we also use the EarlyGlobalRoute tool to perform the subsequent stages (including routing, RC extraction, clock tree synthesis, etc.) and evaluate the PPA metrics, including routed wirelength (wirelength of all modules after routing), routed vertical and horizontal congestion overflow (congestion that exceeds the limits), worst negative slack (the maximum amount by which the timing fails to meet its required constraint), total negative slack (the sum of all negative slacks across all timing paths that fail to meet their constraints), and the number of violation paths (the number of paths that violate timing constraints). Although these metrics are difficult to obtain and expensive to evaluate, they are crucial in chip design and are used to comprehensively assess chip quality.
\subsection{BBO User-Friendly Interfaces}\label{sec:3-5}
Our proposed BBOPlace-Bench has easy-to-use interfaces, making it straightforward to set up both built-in algorithms and user-defined algorithms. A simple example of running CMA-ES under the MGO formulation on the chip case adaptec1 is shown in Code Example~1. It runs the optimization process by setting parameters in lines~5--13 (e.g., \texttt{benchmark} = \texttt{adaptec1}, \texttt{placer} = \texttt{mgo}), initializing an evaluator (in lines~15--20) and optimizer (in lines~22--28), and then iteratively generating and evaluating solutions in lines~30--35. We also provide a visualization interface that conveniently displays the placement of modules, allowing for an intuitive assessment of the results, as shown in Section~\ref{sec:exp-vis}. For reproducibility, we recommend the provided Docker image as the primary setup path for DREAMPlace-related dependencies.
\begin{python}[caption=Run CMA-ES~\cite{cmaes} under the mask-guided optimization formulation on the chip case adaptec1.]
import argparse
import numpy as np
import cma
from src.evaluator import Evaluator

# Set parameters
parser = argparse.ArgumentParser() 
parser.add_argument("--benchmark", type=str, default="adaptec1")
parser.add_argument("--eval_gp_hpwl", action="store_true", default=False)
parser.add_argument("--placer", type=str, choices=["sp", "mgo", "hpo"], default="mgo")
parser.add_argument("--sigma", type=float, default=0.5)
parser.add_argument("--pop_size", type=int, default=20)
parser.add_argument("--seed", type=int, default=42)
args = parser.parse_args() 

# Initialize the evaluator
evaluator = Evaluator(args)
dim = evaluator.n_dim
xl = evaluator.xl.tolist() 
xu = evaluator.xu.tolist()
x0 = np.random.uniform(low=xl, high=xu, size=dim)

# Initialize CMA-ES
cmaes = cma.CMAEvolutionStrategy(
    x0,  
    args.sigma,  
    {'popsize': args.pop_size,
     'bounds': [xl, xu]}
)

# Run CMA-ES for optimization
while not cmaes.stop():
    solutions = cmaes.ask()
    fitness_values = evaluator.evaluate(solutions)
    cmaes.tell(solutions, fitness_values)
    print(f"Generation {cmaes.countiter}: Best fitness = {min(fitness_values):.6f}")
\end{python}

\section{Experiment}
In this section, we first introduce the experimental settings,
including dataset preprocessing and parameter configurations for different BBO algorithms in Section~\ref{sec:exp-settings}. We then present the results on ISPD 2005~\cite{nam2005ispd2005}, comparing the performance of various problem formulations and BBO algorithms in terms of MP-HPWL and GP-HPWL, and comparing them with advanced RL and analytical methods in Section~\ref{sec:exp-ispd}. Subsequently, we report the findings on ICCAD 2015~\cite{iccad15}, analyzing HPWL results and conducting PPA evaluations to assess real-world chip performance in Section~\ref{sec:exp-iccad}. Finally, we show visualization results of chip layouts to offer intuitive insights into the placement solutions generated by various methods in Section~\ref{sec:exp-vis}.
Additionally, we investigate the influence of different numbers of macros (i.e., different search space dimensions) on optimization efficiency and provide an empirical wall-clock time comparison of the computational costs of different algorithms and problem formulations in the supplementary material. To keep the main text concise, we present the core comparative tables and figures in this section, while benchmark instance statistics, case-by-case detailed tables, and supplementary curves are provided in the supplementary material. The code of BBOPlace-Bench is available at \url{https://github.com/lamda-bbo/BBOPlace-Bench}.

\subsection{Experimental Settings}\label{sec:exp-settings}
In the experiments, we compare different BBO algorithms under three problem formulations in BBOPlace-Bench. The main chip cases are ISPD 2005~\cite{nam2005ispd2005} (adaptec and bigblue series) and ICCAD 2015~\cite{iccad15} (superblue series), with detailed per-case statistics (numbers of macros, standard cells, nets, and pins) listed in the supplementary material.

An important setting for BBO in chip placement is the number of macros, which corresponds to the search space dimension for SP and MGO. For ISPD 2005, we use the number of macros specified in the dataset. For ICCAD 2015, because no macros are specified, we default to defining the 512 largest modules (by area) as macros. This is a deterministic, reproducible benchmark construction rule that emphasizes physically large movable blocks when an explicit macro list is unavailable, rather than a claim that this is the only valid industry definition. Because mixed discrete--continuous DREAMPlace hyperparameters are cumbersome for the optimizers used in our HPO experiments without additional mixed-variable machinery, we formulate HPO as a continuous problem: each tuned hyperparameter corresponds to one real-valued coordinate with its own bounds. For categorical DREAMPlace settings, we decode a coordinate by truncating it toward zero and indexing a fixed ordered list of options using modular arithmetic, rather than one-hot encoding or per-category logit coordinates.

For a fair comparison, we keep the hyperparameters of the compared algorithms in the following experiments as consistent as possible, e.g., the population size is set to 50 for most experiments. Our goal is to assess representative algorithms under practically adopted configurations through a shared evaluation protocol---using defaults from implementations (or libraries), settings in official documentation or original papers, and commonly used community settings when the former are unavailable---rather than to perform exhaustive per-algorithm hyperparameter tuning or method-specific over-tuning for peak performance on each case. We introduce the detailed hyperparameters of each algorithm as follows:

\paragraph{SA} We set the initial temperature to 100, with a decay rate of 0.99 and an update frequency of 100 steps. Mutation operators vary for different formulations. For SP, we apply the \textit{inversion} operator, which randomly selects a start point and an end point from a permutation, and then inverts the sequence between them. For MGO, we apply the \textit{shuffle} operator to shuffle the positions of several randomly selected macros. For HPO, we apply the \textit{random resetting} operator to randomly select one element and assign a new value within the search space.

\paragraph{GA} As a population-based search algorithm, GA uses not only mutation to perturb a single solution, but also crossover to recombine two solutions. We use the same mutation operators as SA. For crossover operators, we use \textit{order crossover} for the SP formulation, where two parent permutations exchange segments while preserving the relative order and positions of the remaining elements, and use \textit{uniform crossover} for MGO and HPO formulations, where each element of the offspring solution comes from either parent with equal probability.

\paragraph{CMA-ES} For MGO and HPO, we set the initial values of sigma to 50 and 0.5, respectively, based on the differences in their search space sizes. For other hyperparameters, we use the default settings in pycma\footnote{\url{https://github.com/CMA-ES/pycma}}.

\paragraph{PSO} We integrate PyPop7~\cite{PyPop7}\footnote{\url{https://github.com/Evolutionary-Intelligence/pypop}} to implement PSO. For MGO and HPO formulations, we set the cognitive learning rate, social learning rate, and maximal ratio of velocities to 2.0, 2.0, and 0.2, respectively.

\paragraph{BO} We use BoTorch~\cite{botorch}\footnote{\url{https://github.com/pytorch/botorch}} to implement our Bayesian optimization algorithm, with the default Matern-5/2 kernel function and expected improvement acquisition function. To better handle high-dimensional search spaces, we use BoTorch's dimensionality-scaled prior~\cite{vanilla-0} as our default setting.

Note that for the SP problem formulation, given its permutation search space, we only run SA and GA.
In our experiments, the evaluation budget for optimizing MP-HPWL is set to 10,000, while that for optimizing GP-HPWL is set to 200 because of the significantly longer evaluation time (detailed empirical wall-clock time comparisons are provided in the supplementary material). Unless a table caption states otherwise, each reported mean and standard deviation aggregates five independent random seeds, which is a common setting in chip placement~\cite{lai2022maskplace,macro-regulator}; we pair these summaries with Wilcoxon rank-sum tests on the reported ranks as in the tables. The only exception in the main experiments is GP-HPWL on ICCAD 2015 (Table~\ref{tab:gp-iccad}), where we use ten independent seeds to mitigate concerns that five runs may be too few for highly stochastic BBO under costly GP-HPWL evaluation.

\subsection{Results on ISPD 2005}\label{sec:exp-ispd}
\paragraph{MP-HPWL Comparisons}
We first compare algorithms in BBOPlace-Bench on ISPD cases in terms of MP-HPWL. In addition to these methods, we include three representative RL methods\footnote{The results for MaskPlace come from our own training, while the results for AlphaChip and EfficientPlace come from~\cite{fast-place}.} (i.e., AlphaChip~\cite{nature-graph,alphachip}, MaskPlace~\cite{lai2022maskplace}, and EfficientPlace~\cite{fast-place}) and one state-of-the-art analytical method DREAMPlace~\cite{lin2020dreamplace} for comparison.
We evaluate final placement quality on the same benchmark instances under the same metrics and the benchmark's evaluation pipeline, and we state each method's search depth explicitly. We do not equate or normalize total computational cost across paradigms (e.g., RL offline training versus BBO query counts). Analytical placers and BBO methods solve each case directly; RL methods include a learning phase on those same instances as per their standard reporting. We present these results as contextual reference points rather than as a claim of strict superiority under a unified cross-paradigm compute budget.
The results are shown in Table~\ref{tab:ispd-avg-rank}, where the detailed results on each case are shown in the tables in the supplementary material.

\begin{table}[t!]
\caption{Average rank on the four experimental settings obtained by compared methods on the six cases of ISPD 2005.}
\label{tab:ispd-avg-rank}
\centering
\begin{tabular}{c|c|c|c}
\toprule
Formulation & Algorithm & MP-Rank & GP-Rank \\
\midrule
\multirow{2}{*}{SP}  & SA             & 16   & 14.83 \\
                     & GA             & 15   & 13.67 \\
\midrule
\multirow{5}{*}{MGO} & SA             & 9.67 & 10.5 \\
                     & GA             & 4.83 & 7.83 \\
                     & CMA-ES         & 8    & 10.67 \\
                     & PSO            & 5.83 & 9.33 \\
                     & BO             & 8.33 & 9 \\
\midrule
\multirow{5}{*}{HPO} & SA             & 5.33 & 4.67 \\
                     & GA             & 3.5  & 2    \\
                     & CMA-ES         & 5.67 & 3.17 \\
                     & PSO            & 4.5  & 2.67 \\
                     & BO             & 7.5  & 2.33 \\
\midrule
\multirow{3}{*}{RL} & AlphaChip      & 13.83 & / \\
                    & MaskPlace      & 10.33 & 13  \\
                    & EfficientPlace & 5.5   & 6.17  \\
\midrule
\multirow{1}{*}{Analytical} & DREAMPlace & 11.67 &  9.83 \\
\bottomrule
\end{tabular}
\end{table}

The performance of the SP formulation is limited due to its rapidly growing combinatorial permutation search space, yielding the poorest results with an average rank of 15--16. Ignoring decoding symmetries, the number of distinct sequence pairs scales as $k!\times k!$, which induces enormous combinatorial growth even for moderate $k$. Local perturbations to a permutation can change decoded macro relations in highly non-local ways, which often yields rugged, non-smooth objective landscapes for methods that assume continuity or smooth local structure and increases the difficulty of incremental improvement under a limited evaluation budget. In contrast, MGO and HPO generally rank higher under this benchmark protocol.
Among BBO algorithms, GA often ranks higher than the others: MGO-GA (i.e., using the MGO problem formulation with GA as the optimizer) achieves an average rank of 4.83, while HPO-GA ranks at 3.5, which is consistent with prior results for wire-mask-guided BBO~\cite{wiremask-bbo}. Mutation and crossover allow GA to recombine useful partial assignments in MGO and effective hyperparameter configurations in HPO under the current evaluation budgets. For SP, GA also outperforms SA, but both SP variants remain low-ranked because the sequence-pair search space is much harder. Thus, the observed advantage of GA should be interpreted as formulation- and protocol-dependent, not as universal superiority over other paradigms.
PSO shows competitive performance (average rank 5.83 for the MGO formulation and 4.5 for the HPO formulation), whereas BO struggles in high-dimensional spaces, e.g., resulting in a lower average rank of 8.33 for the MGO formulation where the dimensionality exceeds $1000$.
It is well known that the performance of BO in high-dimensional spaces may need the assistance of additional techniques~\cite{high-dim}.
CMA-ES is a widely used evolution-strategy method for continuous optimization, and we include it under a unified benchmark protocol as a familiar reference baseline. Since MGO uses a high-dimensional discrete grid, this setting is not where CMA-ES is typically recommended: CMA-ES is primarily designed for continuous real-valued search with moderate effective dimensionality and some degree of local smoothness. Therefore, weaker MGO results for CMA-ES should be interpreted as formulation--optimizer mismatch rather than as a general judgment on evolution strategies. At the same time, this finding highlights an opportunity to develop CMA-ES variants that are better aligned with chip placement formulations (e.g., discrete-aware or mixed-variable adaptations).
Notably, top BBO methods, e.g., HPO-GA~(average rank 3.5), HPO-PSO~(average rank 4.5), and MGO-GA~(average rank 4.83), achieve better average ranks than the RL method EfficientPlace~\cite{fast-place}~(average rank 5.5) and the analytical placer DREAMPlace~\cite{lin2020dreamplace}~(average rank 11.67) on these MP-HPWL comparisons, indicating competitive behavior of BBO in this setting. These observations are formulation- and protocol-dependent, and are not intended as universal rankings across all placement workflows.

\paragraph{GP-HPWL Comparisons}
The detailed GP-HPWL results on the ISPD cases are reported in the supplementary material.
Note that AlphaChip~\cite{nature-graph,alphachip} is not included in the table for comparison because data are unavailable.
Overall, the HPO problem formulation attains the highest rankings in this comparison, because it can tune the hyperparameters of the analytical placer DREAMPlace, which simultaneously accounts for both macros and standard cells.
In contrast, the SP and MGO formulations focus exclusively on macros and disregard standard-cell information, thereby limiting their capacity to optimize GP-HPWL.
As can be observed from these detailed results, when combined with different BBO algorithms (SA, GA, CMA-ES, PSO, and BO), HPO achieves substantially higher rankings than both MGO and SP, with HPO-GA attaining the top average rank (2.0). The SP formulation has the lowest rankings, likely due to its less effective search-space design.
Among the BBO variants, GA is the strongest overall in this GP-HPWL comparison, whereas BO ranks second within both MGO and HPO. PSO remains competitive, especially in HPO, but its GP-HPWL ranks are lower than BO's in Table~\ref{tab:ispd-avg-rank}. This pattern is plausible because optimizing the expensive GP-HPWL objective allows much fewer evaluations than optimizing MP-HPWL, making BO's sample efficiency more useful.
Under GP-HPWL evaluation, the top BBO method (i.e., HPO-GA, average rank 2.0) also ranks ahead of the reported RL approach (e.g., EfficientPlace~\cite{fast-place}, average rank 6.17) and analytical baseline (e.g., DREAMPlace~\cite{lin2020dreamplace}, average rank 9.83) in this table, showing competitive behavior on this metric under the current protocol.

\subsection{Results on ICCAD 2015}\label{sec:exp-iccad}

\begin{figure}[t!]\centering
\includegraphics[width=\linewidth]{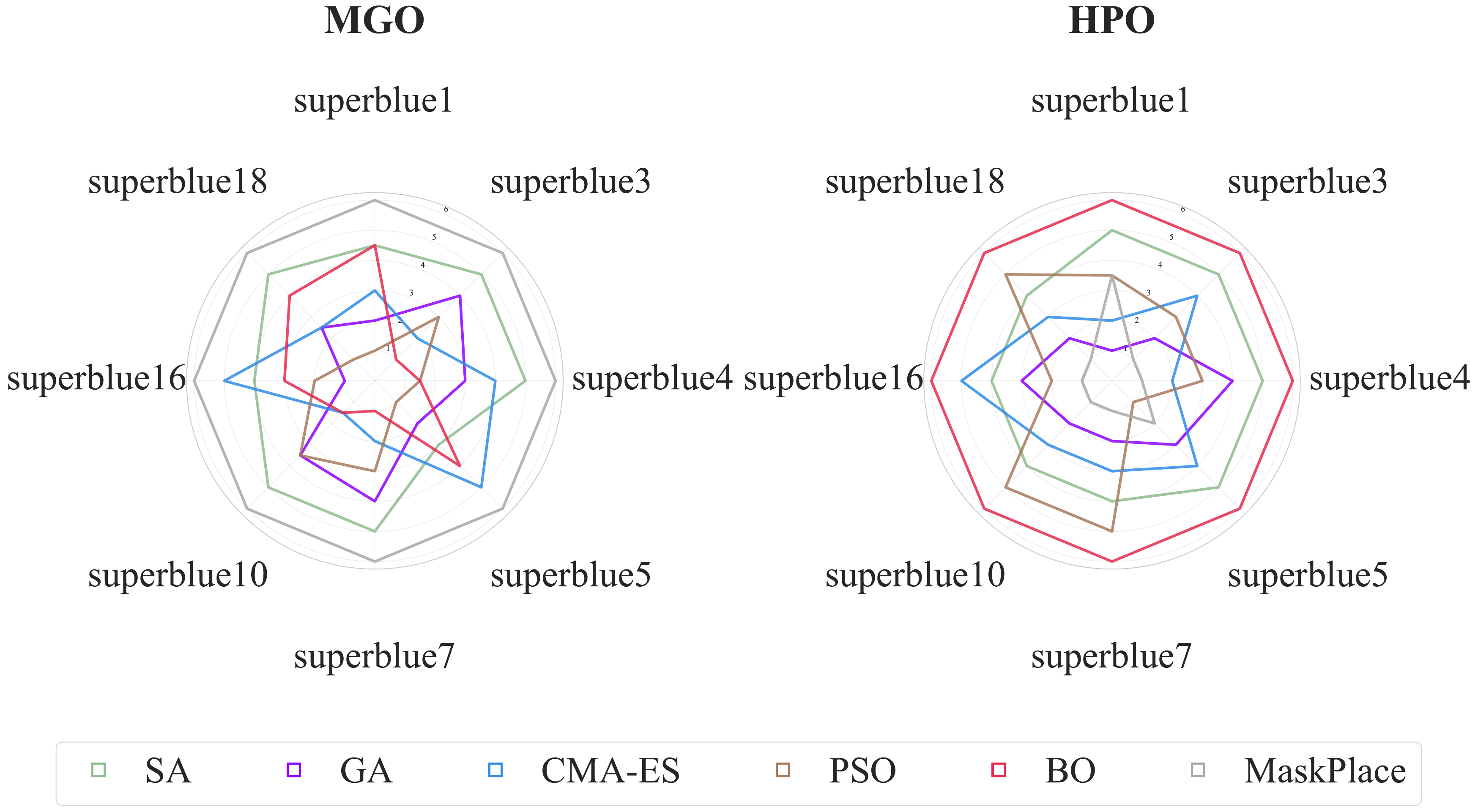}
\caption{Radar plots of average ranks across the ICCAD 2015 cases under MP-HPWL evaluation. Left: MGO formulation. Right: HPO formulation.
}\label{fig:iccad-mp-rader}
\end{figure}

\begin{figure}[t!]\centering
\includegraphics[width=\linewidth]{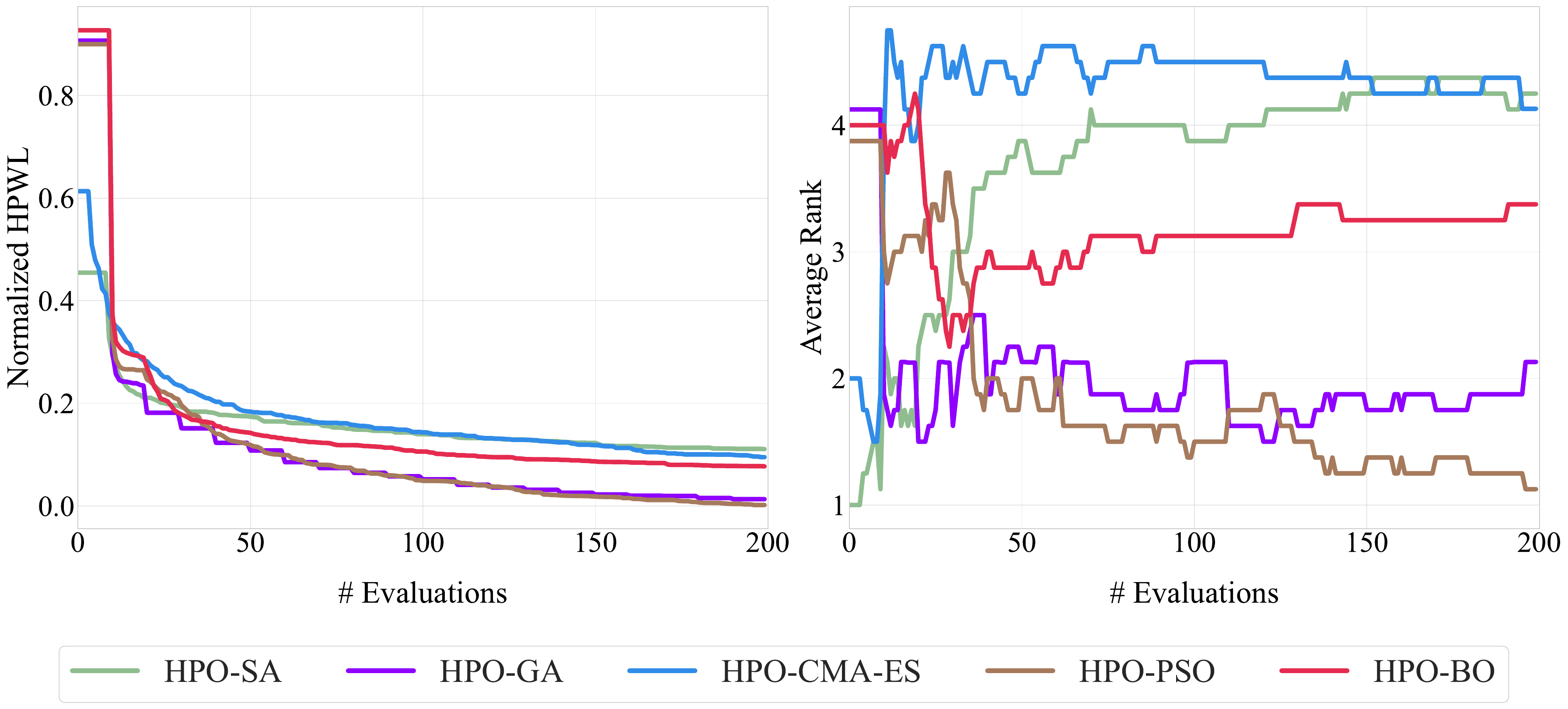}
\caption{Convergence comparison of HPO methods on ICCAD 2015 cases: normalized HPWL and average rank versus \# Evaluations.}

\label{fig:iccad-gp-curve}
\end{figure}

\paragraph{HPWL Comparisons}
The full per-case MP-HPWL numerical results on ICCAD 2015 are reported in the supplementary material, while the detailed GP-HPWL numerical results are reported in Table~\ref{tab:gp-iccad} below; here we further summarize comparative patterns using figures.
Note that AlphaChip~\cite{nature-graph,alphachip} and EfficientPlace~\cite{fast-place} are not included for comparison because data are unavailable.
The GP-HPWL statistics in Table~\ref{tab:gp-iccad} follow the ten-seed protocol described in the experimental settings above.
Our reported results reflect end-to-end \emph{pipeline} behavior: each entry corresponds to a problem formulation (including its decoding or inner placement routines), a black-box outer optimizer, and the shared evaluation chain; they are therefore not intended to isolate optimizer capability from representation and decoding choices. To provide a simple reference under the same MGO decoding and evaluation protocol, we also include random search as an outer-loop baseline so that improvements beyond random exploration are easier to interpret under identical budgets.
For figure-level interpretation, we summarize MP-HPWL average-rank patterns with radar plots in Figure~\ref{fig:iccad-mp-rader}, and show the HPO GP-HPWL convergence trends in Figure~\ref{fig:iccad-gp-curve}; the cross-case GP-HPWL box plots and full broken-axis GP-HPWL curves are provided in the supplementary material.

\begin{table*}[t!]
\caption{GP-HPWL values ($\times 10^7$) obtained by compared methods on the eight cases of ICCAD 2015. Each result consists of the mean and standard deviation of ten runs. The best and runner-up methods on each chip case are \textbf{bolded} and \underline{underlined}, respectively. The symbols `$\approx$' and `-' indicate that the result is almost equivalent and inferior to the best methods, respectively, according to the Wilcoxon rank-sum test with significance level 0.05.}
\label{tab:gp-iccad}
\resizebox{\textwidth}{!}{
\begin{tabular}{c|c|c|c|c|c|c|c|c|c|c}
\toprule
Formulation    & Algorithm & superblue1 & superblue3 & superblue4 & superblue5 & superblue7 & superblue10 & superblue16 & superblue18 & Average Rank \\
\midrule 
\multirow{2}{*}{SP} & SA & 83.45$\pm$3.68 - & 86.94$\pm$4.36 - & 60.15$\pm$2.08 - & 109.15$\pm$5.22 - & 99.99$\pm$1.32 - & 102.07$\pm$2.31 - & 67.16$\pm$1.50 - & 30.04$\pm$0.47 - & 14.25 \\
                    & GA & 80.95$\pm$2.11 - & 81.72$\pm$3.23 - & 56.38$\pm$2.10 - & 100.47$\pm$4.81 - & 96.74$\pm$2.37 - & 98.26$\pm$2.60 - & 65.82$\pm$1.85 - & 29.48$\pm$0.45 - & 13.25 \\
\midrule
\multirow{6}{*}{MGO} & RS & 61.47$\pm$1.15 - & 70.89$\pm$0.92 - & 44.72$\pm$0.11 - & 80.73$\pm$0.43 - & 84.08$\pm$0.60 - & 93.26$\pm$0.91 - & 50.20$\pm$0.49 - & 28.81$\pm$0.13 - & 9.88 \\
                     & SA & 62.22$\pm$1.35 - & 71.89$\pm$1.27 - & 44.78$\pm$0.56 - & 81.65$\pm$1.44 - & 84.74$\pm$1.02 - & 93.67$\pm$0.97 - & 50.44$\pm$0.60 - & 28.81$\pm$0.44 - & 11.13 \\
                     & GA & 55.80$\pm$2.23 - & 62.89$\pm$1.35 - & 40.59$\pm$0.86 - & 74.90$\pm$5.85 - & 76.10$\pm$1.23 - & 88.37$\pm$1.66 - & 46.90$\pm$0.82 - & 26.87$\pm$0.17 - & 6.25 \\
                     & CMA-ES & 60.90$\pm$0.78 - & 71.56$\pm$1.16 - & 44.30$\pm$0.33 - & 80.55$\pm$1.34 - & 83.40$\pm$1.39 - & 93.47$\pm$0.41 - & 50.13$\pm$0.56 - & 28.56$\pm$0.12 - & 9.25 \\
                     & PSO & 57.35$\pm$3.33 - & 66.83$\pm$1.76 - & 41.71$\pm$1.26 - & 78.54$\pm$2.52 - & 79.69$\pm$2.27 - & 89.47$\pm$2.53 - & 47.91$\pm$0.88 - & 27.52$\pm$1.30 - & 7.75 \\
                     & BO & 62.00$\pm$1.27 - & 65.74$\pm$3.33 - & 44.97$\pm$0.54 - & 74.13$\pm$4.13 - & 78.76$\pm$3.13 - & 88.64$\pm$3.10 - & 49.53$\pm$4.49 - & 28.35$\pm$0.95 - & 8.25 \\
\midrule
\multirow{5}{*}{HPO} & SA & 37.41$\pm$0.60 $\approx$ & 42.91$\pm$0.17 - & 28.72$\pm$0.30 - & 40.43$\pm$0.31 $\approx$ & 54.53$\pm$0.49 - & 69.25$\pm$0.40 - & 37.02$\pm$0.31 $\approx$ & 22.23$\pm$0.22 - & 4.25 \\
                     & GA & 37.13$\pm$0.81 $\approx$ & \underline{42.34$\pm$0.34 $\approx$} & \underline{28.08$\pm$0.26 $\approx$} & \underline{40.20$\pm$0.26 $\approx$} & \underline{53.56$\pm$0.55 $\approx$} & \underline{68.68$\pm$0.30 $\approx$} & \underline{36.85$\pm$0.33 $\approx$} & \underline{21.96$\pm$0.12 $\approx$} & 2.13 \\
                     & CMA-ES & 37.95$\pm$0.41 - & 42.77$\pm$0.40 $\approx$ & 28.39$\pm$0.23 - & 40.46$\pm$0.34 $\approx$ & 54.63$\pm$0.62 - & 69.29$\pm$0.86 $\approx$ & 37.00$\pm$0.20 - & 22.01$\pm$0.10 $\approx$ & 4.13 \\
                     & PSO & \underline{37.09$\pm$0.63 $\approx$} & \textbf{42.31$\pm$0.47 $\approx$} & \textbf{28.07$\pm$0.11 $\approx$} & \textbf{40.11$\pm$0.21 $\approx$} & \textbf{53.32$\pm$0.42 $\approx$} & \textbf{68.57$\pm$0.25 $\approx$} & \textbf{36.65$\pm$0.25 $\approx$} & \textbf{21.88$\pm$0.12 $\approx$} & 1.13 \\
                     & BO & \textbf{37.00$\pm$0.56 $\approx$} & 43.90$\pm$1.31 - & 28.26$\pm$0.19 $\approx$ & 40.34$\pm$0.19 $\approx$ & 53.90$\pm$0.80 $\approx$ & 69.40$\pm$0.83 $\approx$ & 36.90$\pm$0.28 $\approx$ & 22.11$\pm$0.09 - & 3.38 \\
\midrule
\multirow{1}{*}{RL} & MaskPlace & 73.59$\pm$1.68 - & 81.12$\pm$4.57 - & 50.59$\pm$3.49 - & 91.50$\pm$3.00 - & 95.77$\pm$0.43 - & 102.89$\pm$4.75 - & 55.12$\pm$0.96 - & 32.74$\pm$0.90 - & 13 \\
\midrule
\multirow{1}{*}{Analytical} & DREAMPlace & 69.92$\pm$2.73 - & 88.53$\pm$3.18 - & 44.58$\pm$5.21 - & 77.28$\pm$5.00 - & 107.20$\pm$2.17 - & 110.60$\pm$4.54 - & 67.58$\pm$4.50 - & 25.68$\pm$0.28 - & 11.88 \\
\bottomrule
\end{tabular}
}

\end{table*}

The detailed MP-HPWL results in the supplementary material show that the MGO formulation ranks highest on MP-HPWL in this benchmark setting, likely due to its advantage in the dataset’s low-dimensional search space (fixed 512 macros), where MGO-PSO achieves the lowest MP-HPWL with the best average rank of 1.88,
followed by MGO-BO with an average rank of 2.38.
HPO variants rank lower than MGO on MP-HPWL, and the SP formulation remains the lowest in these MP-HPWL comparisons.
Similar to the GP-HPWL results on ISPD, Table~\ref{tab:gp-iccad} shows that the HPO formulation continues to rank higher in GP-HPWL on ICCAD.
To further illustrate the performance of different HPO variants, we plot the convergence curve of GP-HPWL under the HPO formulation, as shown in Figure~\ref{fig:iccad-gp-curve}.
Figure~\ref{fig:iccad-gp-curve} focuses on HPO variants: HPO-GA and HPO-PSO reduce normalized HPWL quickly and maintain stronger average-rank trajectories by the end of the 200 evaluations, whereas HPO-CMA-ES and HPO-BO plateau at higher ranks. The comparisons with MaskPlace~\cite{lai2022maskplace} and DREAMPlace~\cite{lin2020dreamplace} are reported in Table~\ref{tab:gp-iccad}.
Compared with DREAMPlace's mixed placement, which places macros and standard cells simultaneously in a single stage, our HPO formulation under GP-HPWL evaluation adopts a two-stage placement and often shows an earlier decrease in GP-HPWL by considering the critical macros first: it first places the macros by running DREAMPlace, and then places the standard cells by running DREAMPlace again to perform GP-HPWL evaluation.
MaskPlace has the weakest average rank on the GP-HPWL metric in this comparison, likely because it only considers macro information.
The performance of BO is worse than that of GA and PSO on most cases, regardless of whether the MGO or HPO formulation is used.
This trend contradicts the common expectation that BO outperforms GA in sample efficiency, which underscores the need for more targeted chip-placement techniques that take advantage of BO's ability to effectively use past evaluations.

\begin{table*}[t!]
\caption{PPA metrics obtained by compared methods under the HPO formulation setup on ICCAD 2015 benchmarks. Among these metrics, rWL (m) is the routed wirelength, rO-H (\%) and rO-V (\%) represent the routed horizontal and vertical congestion overflow, respectively, WNS (ns) is the worst negative slack, TNS (1e5$\mu$s) is the total negative slack, and NVP (1e4) is the number of violation paths. The best and runner-up methods are \textbf{bolded} and \underline{underlined}, respectively.}
\label{tab:ppa1}
\centering
\begin{tabular}{c|c|c|c|c|c|c|c|c}
\toprule
Benchmark    & Formulation & Algorithm & rWL (m) $\downarrow$& rO-H (\%) $\downarrow$& rO-V (\%) $\downarrow$& WNS (ns) $\uparrow$& TNS (1e5 $\mu$s) $\uparrow$& NVP (1e4) $\downarrow$\\
\midrule
\multirow{5}{*}{superblue1} 
                            & \multirow{5}{*}{HPO} & SA & \textbfu{75.06} & \textbf{0.00}  & \textbf{0.00}  & -21.23  & -0.16 & \textbfu{0.60} \\
                            &                      & GA	& \textbf{74.90} & \textbf{0.00} & \textbf{0.00}  & -21.82 & \textbf{-0.10} & \textbf{0.50} \\
                            &                      & CMA-ES & 77.55 & \textbf{0.00} & \textbf{0.00}  & \textbfu{-20.35} & -0.19 & 0.73 \\
                                                        &                      & PSO& 75.23 & \textbfu{0.01} & \textbf{0.00} & -25.04 & \textbfu{-0.11} & \textbf{0.50} \\ 

                            &                      & BO & 80.33 & 0.08  &\textbfu{0.02}  & \textbf{-16.91}  & -0.16 & 0.75 \\ 
\midrule
\multirow{5}{*}{superblue3} 
                            & \multirow{5}{*}{HPO} & SA & 88.51 & 0.29 & \underline{0.01}  & \textbf{-23.77} & \textbf{-0.11} & \underline{0.35} \\
                            &                      & GA	& \textbfu{86.66} & 0.13 & \underline{0.01}  & \underline{-23.85} & \textbf{-0.11} & \underline{0.35} \\
                            &                      & CMA-ES & 87.97 & \textbf{0.04} & \textbf{0.00}  & -26.70 & \underline{-0.13} & \textbf{0.34} \\
                            &                      & PSO& \textbf{86.41} & \textbfu{0.12} & \textbfu{0.01} & -26.63 & \textbf{-0.11} & 0.38 \\
                            &                      & BO & 91.24 & 0.44 & \underline{0.01}  & -27.32  & -0.17 & 0.46 \\
\midrule
\multirow{5}{*}{superblue4} 
                            & \multirow{5}{*}{HPO} & SA & 58.61 & 0.04  & \underline{0.01}  & -25.42  & \textbfu{-0.28} & \textbfu{0.59} \\
                            &                      & GA	& \textbfu{57.97} & \underline{0.03} & \textbf{0.00}  & \textbfu{-19.17}  & -0.29 & 0.61 \\
                            &                      & CMA-ES & 58.60 & \underline{0.03} & 0.03  & -26.96  & -0.29 & 0.60 \\
                            &                      & PSO& \textbf{57.51} & 0.05 & \textbf{0.00} & \textbf{-15.04} & \textbf{-0.26} & \textbf{0.58} \\
                            &                      & BO & 59.23 & \textbf{0.02} & \textbf{0.00}  & -19.34  & -0.30 & 0.61 \\
\midrule
\multirow{5}{*}{superblue5} 
                            & \multirow{5}{*}{HPO} & SA & 82.58 & \underline{0.11}  & \underline{0.02}	 & -60.39  & -0.17 & \textbfu{0.45} \\
                            &                      & GA	& \textbf{81.42} & 0.15  & \textbf{0.01}  & -46.08  & \textbf{-0.14} & 0.51 \\
                            &                      & CMA-ES & 82.90 & 0.26 & \underline{0.02}  & \underline{-45.90}  & -0.17 & 0.57 \\
                            &                      & PSO& \textbfu{81.43} & 0.29 & \textbf{0.01} & -52.35 & \textbfu{-0.16} & \textbf{0.42} \\
                            &                      & BO & 84.85 & \textbf{0.10}  & \textbf{0.01}  & \textbf{-44.16} & -0.18 & 0.50 \\ 
\midrule
\multirow{5}{*}{superblue7} 
                            & \multirow{5}{*}{HPO} & SA & 116.37 & \underline{0.01} & \textbf{0.00} & -13.81 & \underline{-0.14} & 0.81  \\
                            &                      & GA	& \textbfu{113.17} & \underline{0.01} & \textbf{0.00} & -13.88 & \textbf{-0.12} & \textbf{0.71}  \\
                            &                      & CMA-ES & 115.10 & \textbf{0.00} & \textbf{0.00} & -13.26 & \underline{-0.14} & 0.81  \\
                            &                      & PSO& \textbf{111.84} & \textbf{0.00} & \textbf{0.00} & \textbf{-11.90} & \textbf{-0.12} & \textbfu{0.80} \\
                            &                      & BO & 116.68 & \underline{0.01} & \textbf{0.00} & \textbfu{-12.30} & \underline{-0.14} & 0.84  \\ 
\midrule
\multirow{5}{*}{superblue10}
                            & \multirow{5}{*}{HPO} & SA & 140.73 & \textbf{0.01} & \textbf{0.00} & -26.93 & \underline{-0.77} & \textbf{1.12}  \\
                            &                      & GA	& \textbfu{139.07} & \textbf{0.01} & \textbf{0.00} & \underline{-22.70} & -0.78 & 1.20  \\
                            &                      & CMA-ES & 140.56 & \underline{0.02} & \textbf{0.00} & \textbf{-20.28} & \textbf{-0.66} & \underline{1.15}  \\
                            &                      & PSO& \textbf{138.67} & 0.06 & \textbfu{0.01} & -25.62 & -0.91 & 1.29 \\
                            &                      & BO & 140.35 & \textbf{0.01} & \textbf{0.00} & -24.61 & -0.88 & 1.29  \\
\midrule
\multirow{5}{*}{superblue16}
                            & \multirow{5}{*}{HPO} & SA &  74.99 & \textbf{0.03} & \textbf{0.00} & \underline{-16.08} & \textbf{-0.30} & \textbf{1.33} \\
                            &                      & GA	&  74.23 & 0.21 & \underline{0.01} & -16.37 & -0.36 & 1.52 \\
                            &                      & CMA-ES &  \textbfu{74.00} & 0.16 & \underline{0.01} & -16.14 & \underline{-0.31} & \underline{1.44} \\
                            &                      & PSO&  \textbf{73.52} & 0.17 & \textbfu{0.01} & -16.55 & -0.32 & 1.55 \\
                            &                      & BO &  75.02 & \underline{0.06} & \textbf{0.00} & \textbf{-15.27} & \textbf{-0.30} & \textbf{1.33} \\
\midrule
\multirow{5}{*}{superblue18}
                            & \multirow{5}{*}{HPO} & SA &  46.44 & \textbf{0.02} & \textbf{0.00} & \textbf{-8.46} & \textbf{-0.04} & \textbf{0.21}  \\
                            &                      & GA	&  \textbf{45.36} & 0.07 & \underline{0.01} & \underline{-10.58} & \underline{-0.07} & 0.32 \\
                            &                      & CMA-ES &  45.96 & \underline{0.05} & \underline{0.01} & -13.73 & \underline{-0.07} & \underline{0.28} \\
                            &                      & PSO&  \textbfu{45.57} & \textbfu{0.04} & \textbf{0.00} & -12.70 & -0.15 & 0.53 \\
                            &                      & BO &  46.44 & 0.09 & \underline{0.01} & -12.75 & -0.09 & 0.30 \\
\bottomrule
\end{tabular}
\end{table*}

\paragraph{PPA Comparisons}
Since GP-HPWL is more closely aligned with the final chip performance and the HPO formulation shows stronger GP-HPWL rankings on the ICCAD 2015 dataset, we focus our PPA evaluations exclusively on the solutions derived from the HPO formulation when optimizing GP-HPWL to assess real-world manufacturing metrics. For each method, we select the best chip placement result from multiple runs based on GP-HPWL for PPA evaluation.
The PPA evaluation results for the HPO formulation are shown in Table~\ref{tab:ppa1}, where rWL (m) is the routed wirelength, rO-H (\%) and rO-V (\%) represent the routed horizontal and vertical congestion overflow, respectively, WNS (ns) is the worst negative slack, TNS (1e5$\mu$s) is the total negative slack, and NVP (1e4) is the number of violation paths.
For WNS and TNS, larger values are better, while for the other metrics, smaller values are better.
As shown in Table~\ref{tab:ppa1}, no single BBO algorithm dominates across all performance metrics, which highlights the complexity of optimizing real-world chip performance.
Similar to the GP-HPWL results in Table~\ref{tab:gp-iccad}, PSO and GA generally yield relatively lower routed wirelength outcomes: PSO ranks first in five cases, while GA does so in three cases. The values of rO-H and rO-V are all very small ($\leq$ 0.3\% and $\leq$ 0.03\%, respectively), suggesting that all algorithms keep congestion relatively low. WNS measures the most timing-critical path in the chip, while TNS measures the overall severity of timing issues.
For a clearer analysis, we calculate the Pearson correlation coefficients: the correlation between rWL and WNS is -0.10, and that between rWL and TNS is -0.65. The value of -0.10 implies a weak relationship between wirelength and WNS. For instance, SA and BO, despite having less favorable GP-HPWL results, achieve the best WNS in two and three cases, respectively.
As for TNS, GA and PSO have the best results, which are closely related to their excellent performance in rWL. The strong correlation between NVP and TNS (with Pearson correlation coefficient -0.68) indicates that fewer violation paths lead to better TNS, which is consistent with previous research~\cite{macro-regulator}. In summary, these results suggest that for chip placement, BBO needs further progress in surrogate modeling and multi-objective optimization to meet PPA requirements.

\subsection{Visualization Analysis}\label{sec:exp-vis}
\begin{figure*}[t!]
\centering
\includegraphics[width=0.24\textwidth]{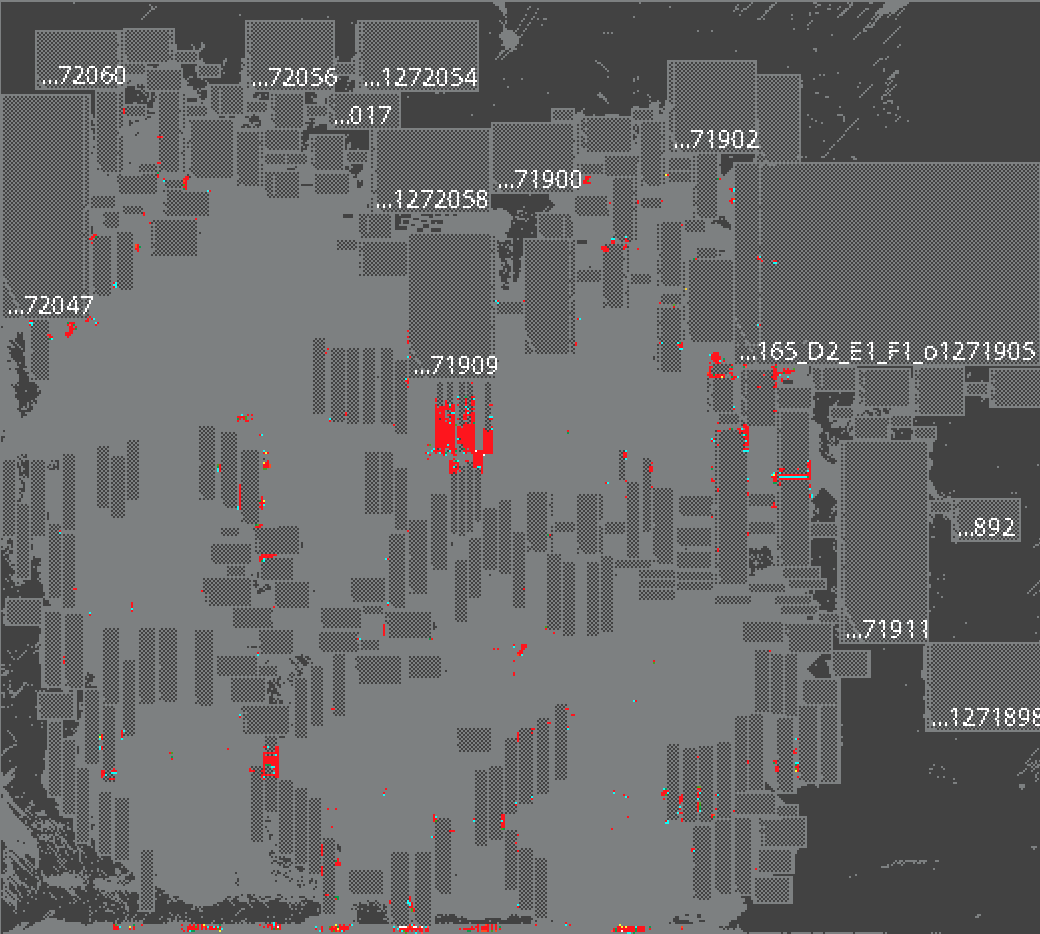}
\includegraphics[width=0.24\textwidth]{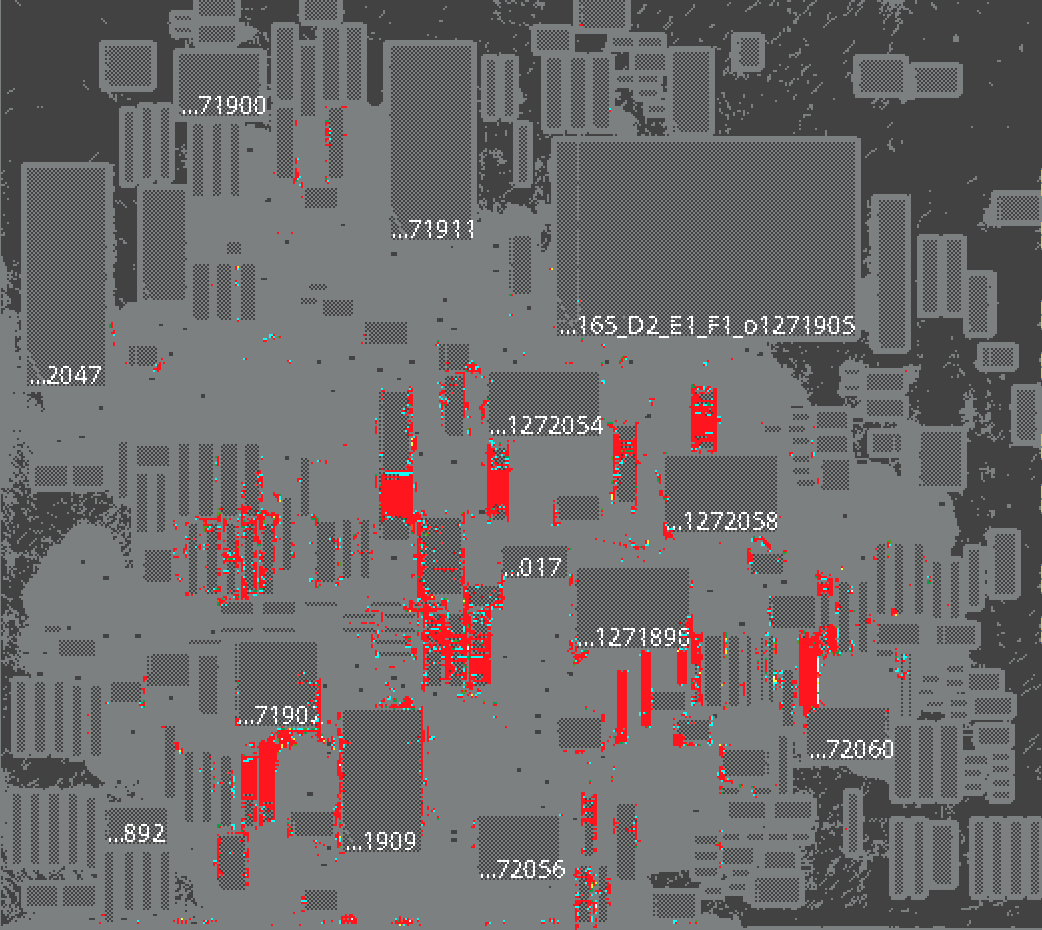}
\includegraphics[width=0.24\textwidth]{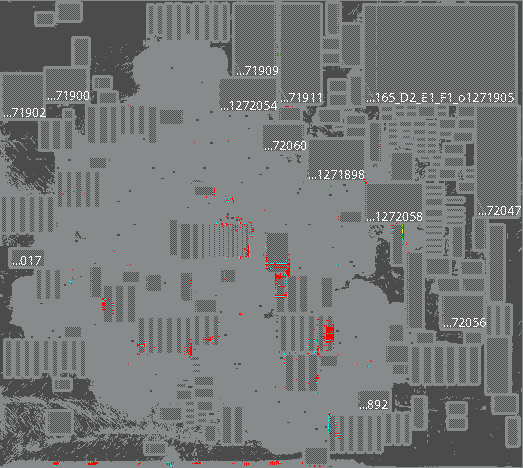}
\includegraphics[width=0.24\textwidth]{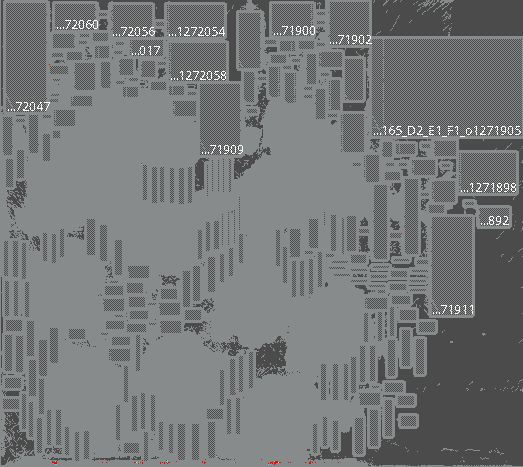}
 \\
\begin{minipage}{0.24\textwidth}
\centering
(a) DREAMPlace~\cite{lin2020dreamplace}
\end{minipage}
\begin{minipage}{0.24\textwidth}
\centering
(b) MaskPlace~\cite{lai2022maskplace}
\end{minipage}
\begin{minipage}{0.24\textwidth}
\centering
(c) MGO-GA
\end{minipage}
\begin{minipage}{0.24\textwidth}
\centering
(d) HPO-PSO
\end{minipage} \\
\caption{Placement layouts and congestion maps of different methods on the chip case superblue7 of ICCAD 2015. The congestion results are obtained by the EarlyGlobalRoute tool, where the red points indicate congestion-critical regions.}
\label{fig:vis-superblue7}
\end{figure*}

To provide an intuitive comparison, we present the visualization results of various methods evaluated on ICCAD 2015, as shown in Figure~\ref{fig:vis-superblue7}. Our proposed BBOPlace-Bench provides a convenient visualization interface that assists users unfamiliar with chip placement in understanding the outputs of their BBO algorithms. Compared with the advanced analytical placer DREAMPlace~\cite{lin2020dreamplace} and the RL placer MaskPlace~\cite{lai2022maskplace}, BBO methods equipped with different problem formulations and optimization algorithms show different placement patterns, and several BBO configurations exhibit fewer critical congestion points in this visualization.

\section{Conclusions and Discussions}
In this paper, we propose BBOPlace-Bench, which is the first BBO benchmark for chip placement.
BBOPlace-Bench decouples problem formulation, optimization algorithm, and evaluation, offering a flexible framework that allows users to easily implement and test their BBO algorithms. The framework is intended to facilitate the application of BBO and help address the significant problem of chip placement. One limitation of this paper is that we used the EarlyGlobalRoute software tool for PPA evaluation. We plan to integrate open-source EDA tools (e.g., OpenROAD~\cite{openroad} and iEDA~\cite{ieda}) into our benchmark to facilitate comprehensive performance evaluation in fully open-source workflows. We also plan to incorporate more advanced chip designs, which can broaden the set of available test cases. One interesting future direction is to apply fitness landscape analysis~\cite{ela,ela-1,ela-2} to unravel the intrinsic properties of chip placement problems, which may deepen our understanding and thereby guide the development of more targeted BBO algorithms tailored to address the specific challenges of this domain.

Based on our empirical findings, several promising directions emerge for advancing BBO for chip placement, each addressing critical challenges in the field:

\begin{itemize}

\item \textbf{Multi-objective optimization.} Beyond the wirelength minimization explored in this work, chip design requires the simultaneous optimization of several possibly conflicting objectives~\cite{mo-place,efficient-tdp}, including routing congestion, dynamic power consumption, timing slack, and area utilization, all of which collectively determine the final chip quality. Applying and developing multi-objective BBO algorithms~\cite{moea-book,large-scale-mo-book,offline-moo} that can balance these trade-offs while maintaining computational efficiency remains a pressing need for chip placement.

\item \textbf{Expensive optimization.} Many real-world applications of BBO involve expensive objective function evaluations, with chip design serving as a representative example. Developing accurate surrogate models~\cite{data-driven-ea,two-arch} for GP-HPWL evaluation or even PPA metrics~\cite{zheng2025deepgate4,lamplace} to reduce reliance on expensive full evaluations is a promising future direction. Advanced algorithms for expensive optimization, such as trust-region search~\cite{turbo}, could also be considered.

\item \textbf{High-dimensional optimization.} Chip placement inherently involves high-dimensional search spaces (e.g., coordinates and orientations for thousands of macros).
Besides applying general dimensionality reduction techniques~\cite{high-dim,large-scale-metaheuristics,mctsvs}, future work could focus on methods that exploit problem-specific structures to mitigate the high-dimensional challenge, such as macro groupings and hierarchical dependencies between placement stages.

\item \textbf{Constrained optimization.} Extending BBOPlace-Bench to explicitly support constrained optimization~\cite{constrain-0,constrain-1} is a promising direction. Currently, constraints (e.g., non-overlapping) are handled implicitly through the decoding mechanisms (e.g., MGO) or the underlying placer (e.g., HPO). Exposing explicit constraint violation metrics (e.g., density or routing congestion violations) would allow the evaluation of advanced constrained BBO algorithms, which is crucial for real-world chip design.

\item \textbf{Learning-enhanced optimization.} Leveraging learning approaches to automate BBO algorithms~\cite{metabbo-survey}, such as optimization by learning from offline datasets~\cite{design-bench,soobench}, automated operator selection and configuration~\cite{madac,ma2023metabox,guo2025configx}, or automated BBO algorithm design~\cite{llm4ad-survey,EoH,llamea}, could enable robust algorithms that generalize across diverse placement cases.
\end{itemize}

\clearpage
\appendices
\section{Additional Results}

\subsection{Benchmark instance statistics}\label{sec:chip-stats}
This subsection lists the numbers of macros, standard cells, nets, and pins for each benchmark chip used in the experiments.
\begin{table}[ht]
\caption{Detailed statistics of the chips, where \# Macros, \# Cells, \# Nets, and \# Pins denote the number of macros, standard cells, nets, and pins contained by a chip, respectively.}\label{tab:statistics}
\centering
\begin{tabular}{@{}l|rrrr@{}}
\toprule
Chips   & \# Macros & \# Cells   & \# Nets    & \# Pins    \\ \midrule 
adaptec1    & 543       & 210,904   & 221,142   & 944,053   \\
adaptec2    & 566       & 254,457   & 266,009   & 1,069,482 \\
adaptec3    & 723       & 450,927   & 466,758   & 1,875,039 \\
adaptec4    & 1,329      & 494,716   & 515,951   & 1,912,420 \\
bigblue1    & 560       & 277,604   & 284,479   & 1,144,691 \\
bigblue3    & 1,298      & 1,095,514 & 1,123,170 & 3,833,218 \\
\midrule
superblue1  & 512       & 1,209,716 & 1215710   & 3,767,494 \\
superblue3  & 512       & 1,213,253 & 1,224,979 & 3,905,321 \\
superblue4  & 512       & 795,645   & 802,513   & 2,497,940 \\
superblue5  & 512       & 1,086,888 & 1,100,825 & 3,246,878 \\
superblue7  & 512       & 1,931,639 & 1,933,945 & 6,372,094 \\
superblue10 & 512       & 1,876,103 & 1,898,119 & 5,560,506 \\
superblue16 & 512       & 981,559   & 999,902   & 3,013,268 \\
superblue18 & 512       & 768,068   & 771,542   & 2,559,143 \\
\bottomrule
\end{tabular}
\end{table}

\subsection{Influence of Different Numbers of Macros}\label{sec:exp-sensitivity}

\begin{table*}[b!]
\caption{Influence analysis of the number of macros: GP-HPWL values ($\times 10^7$) obtained by the two best-performing methods (i.e., GA and PSO) on ICCAD 2015 under the MGO formulation. Each result consists of the mean and standard deviation of five runs. The best and runner-up settings are \textbf{bolded} and \underline{underlined}, respectively.}\label{tab:gp-n-macro}
\centering
\resizebox{\textwidth}{!}{
\begin{tabular}{c|c|c|c|c|c|c|c|c|c}
\toprule
Algorithm & Number of macros & superblue1 & superblue3 & superblue4 & superblue5 & superblue7 & superblue10 & superblue16 & superblue18 \\ 
\midrule
\multirow{4}{*}{MGO-GA} & 128    & \textbf{52.07$\pm$0.37} & \textbf{59.89$\pm$0.79} & \textbfu{37.81$\pm$0.32} & 
                    \textbf{64.08$\pm$1.18} & \textbf{70.54$\pm$1.72} & \textbf{87.89$\pm$1.39} & \textbfu{46.08$\pm$0.42} & \textbf{25.72$\pm$0.09} \\
                    & 256    & 53.91$\pm$0.52 & \textbfu{61.74$\pm$0.14} & 39.04$\pm$0.33 & 70.66$\pm$0.33 & 75.83$\pm$1.58 & 88.73$\pm$0.52 & \textbf{45.99$\pm$0.72} & \textbfu{25.96$\pm$0.24} \\
                    & 512    & 55.35$\pm$1.24 & 62.62$\pm$1.24 & 40.24$\pm$0.65 & 72.15$\pm$1.16 & 75.79$\pm$1.47 & 88.69$\pm$1.35 & 46.71$\pm$0.77 & 26.96$\pm$0.15 \\
                    & 1024   & 60.27$\pm$0.68 & 68.29$\pm$0.48 & 45.77$\pm$0.27 & 72.99$\pm$1.00 & 78.70$\pm$1.06 & 89.72$\pm$1.26 & 48.33$\pm$1.28 & 27.59$\pm$0.20 \\
\midrule
\multirow{4}{*}{MGO-PSO}& 128    & \textbfu{52.86$\pm$0.17} & 63.31$\pm$0.59 & \textbf{36.73$\pm$0.04} & 
                    \textbfu{65.57$\pm$4.47} & \textbfu{72.29$\pm$1.67} & 89.76$\pm$1.33 & 46.87$\pm$1.07 & 26.93$\pm$0.26 \\       
                    & 256    & 55.58$\pm$0.43 & 66.41$\pm$1.37 & 38.91$\pm$0.03 & 77.99$\pm$2.84 & 83.04$\pm$1.63 & \textbfu{88.56$\pm$3.20} & 46.19$\pm$0.14 & 26.76$\pm$0.18 \\
                    & 512    & 56.65$\pm$1.15 & 66.53$\pm$1.48 & 41.09$\pm$0.73 & 78.00$\pm$3.13 & 79.91$\pm$2.74 & 90.90$\pm$2.04 & 47.96$\pm$0.62 & 26.98$\pm$0.11 \\
                    & 1024   & 63.84$\pm$0.51 & 72.93$\pm$2.02 & 44.73$\pm$0.21 & 78.88$\pm$0.78 & 85.17$\pm$0.96 & 88.74$\pm$3.06 & 48.41$\pm$0.59 & 27.72$\pm$0.09 \\
\bottomrule
\end{tabular}}
\end{table*}

As we mentioned before, an important setting in the MGO formulation is the number of macros, which defines the problem dimension, because a solution under the MGO formulation records the coordinates of all macros directly. We examine this setting on ICCAD 2015 designs, selecting the two best algorithms (GA and PSO) to assess the influence of the number of macros.
As shown in Table~\ref{tab:gp-n-macro}, the overall results deteriorate as the number of macros increases, because the optimization difficulty increases with the number of macros, while the number of evaluations is kept fixed.
When the number of macros is small, the MGO formulation prioritizes the more important macros (i.e., those with larger connected areas) under a limited evaluation budget, yielding better GP-HPWL values.
In certain instances, such as superblue10 and superblue16, adjusting a greater number of macros occasionally yields superior results, which may be because the given evaluation budget is still sufficient for the enlarged search space.
Overall, our BBOPlace-Bench offers flexible problem formulation choices, providing researchers with a controllable test-bed for investigating high-dimensional BBO challenges.

\subsection{Empirical Wall-Clock Time Comparison}\label{sec:exp-runtime}
Due to the differences in problem formulation and optimization algorithms, the runtime varies significantly across different methods. Here, we present the average per-iteration runtime on the chip case adaptec3, as shown in Table~\ref{tab:run-time}. The runtime is partitioned into two components: Optimization time is the runtime of algorithm execution, and evaluation time is the runtime of MP-HPWL or GP-HPWL evaluation. Because the per-iteration optimization time of BO and CMA-ES increases with the number of iterations, we average it over all completed iterations for each algorithm. The per-iteration optimization time of BO ($\approx$1--9~s) and CMA-ES ($\approx$0.2--1.0~s) is one to two orders of magnitude higher than that of GA ($\approx$0.03--0.7~s) and SA ($\approx$0.01--0.7~s), reflecting the cubic complexity of Gaussian process training and inference in BO and the covariance-matrix adaptation overhead in CMA-ES. PSO exhibits the lowest optimization cost ($<$0.04~s) owing to its lightweight velocity updates.

\begin{table}[htbp]
\caption{Optimization and evaluation time of each iteration of different methods on the chip case adaptec3, where the optimization time is the wall-clock time of algorithm execution and the evaluation time is the wall-clock time of MP-HPWL evaluation or GP-HPWL evaluation.}\label{tab:run-time}
\centering
\resizebox{0.48\textwidth}{!}{
\begin{tabular}{c|c|c|c|c}
\toprule
\multicolumn{3}{c|}{Evaluation Settings}  & Optimization Time & Evaluation Time \\ 
\midrule
\multirow{12}{*}{MP-HPWL} & \multirow{2}{*}{SP}   & SA                   & 0.0271               & \multirow{2}{*}{0.3870}         \\
                     &                       & GA                   & 0.0296               &                                  \\\cmidrule{2-5}
                     & \multirow{5}{*}{MGO} & SA                   & 0.0514               & \multirow{5}{*}{2.5114}        \\
                     &                       & GA                   & 0.1751               &                                  \\
                                          &                       & CMA-ES             & 1.0171               &                                  \\
                     &                       & PSO                  & 0.0021               &                                  \\
                     &                       & BO                   & 5.3156               &                                  \\
                     \cmidrule{2-5}
                     & \multirow{5}{*}{HPO}  & SA                   & 0.0182               & \multirow{5}{*}{62.9666}       \\
                     &                       & GA                   & 0.0494               &                                  \\
                    &                       & CMA-ES              & 0.3777               &                                  \\
                    &                       & PSO                  & 0.0118               &                                  \\
                     &                       & BO                   & 1.1098               &                                  \\
\midrule
\multirow{12}{*}{GP-HPWL} & \multirow{2}{*}{SP}   & SA                   & 0.1496               & \multirow{2}{*}{18.5297}        \\
                     &                       & GA                   & 0.6337               &                                  \\\cmidrule{2-5}
                     & \multirow{5}{*}{MGO} & SA                   & 0.7083               & \multirow{5}{*}{20.3188}        \\
                     &                       & GA                   & 0.6855               &                                  \\
                                          &                       & CMA-ES              & 1.0578               &                                  \\
                     &                       & PSO                  & 0.0390               &                                  \\
                     &                       & BO                   & 2.6328               &                                  \\
                     \cmidrule{2-5}
                     & \multirow{5}{*}{HPO}  & SA                   & 0.1755               & \multirow{5}{*}{81.5007}       \\
                     &                       & GA                   & 0.4483               &                                  \\
                                          &                       & CMA-ES             & 0.2507               &                                  \\
                     &                       & PSO                  & 0.0028               &                                  \\
                     &                       & BO                   & 9.1910               &                                  \\
\bottomrule
\end{tabular}}
\end{table}

For evaluation time, the SP and MGO formulations require significantly less time under MP-HPWL evaluation compared to the HPO formulation, because of the significant overhead of DREAMPlace for the placement of all modules in the HPO formulation.
The evaluation time of SP and MGO increases considerably under GP-HPWL evaluation, because it requires DREAMPlace to place standard cells following the macro placement stage.
Nevertheless, the resulting evaluation time of SP and MGO formulations remains lower than that of the HPO formulation, since their macro placement steps are inherently faster.
BBOPlace-Bench is designed to support flexible problem formulations and evaluation settings, allowing researchers to examine BBO algorithms at different computational costs.

\subsection{Detailed results}\label{sec:exp-detailed-results}

\begin{figure*}[t!]\centering
\includegraphics[width=0.99\linewidth]{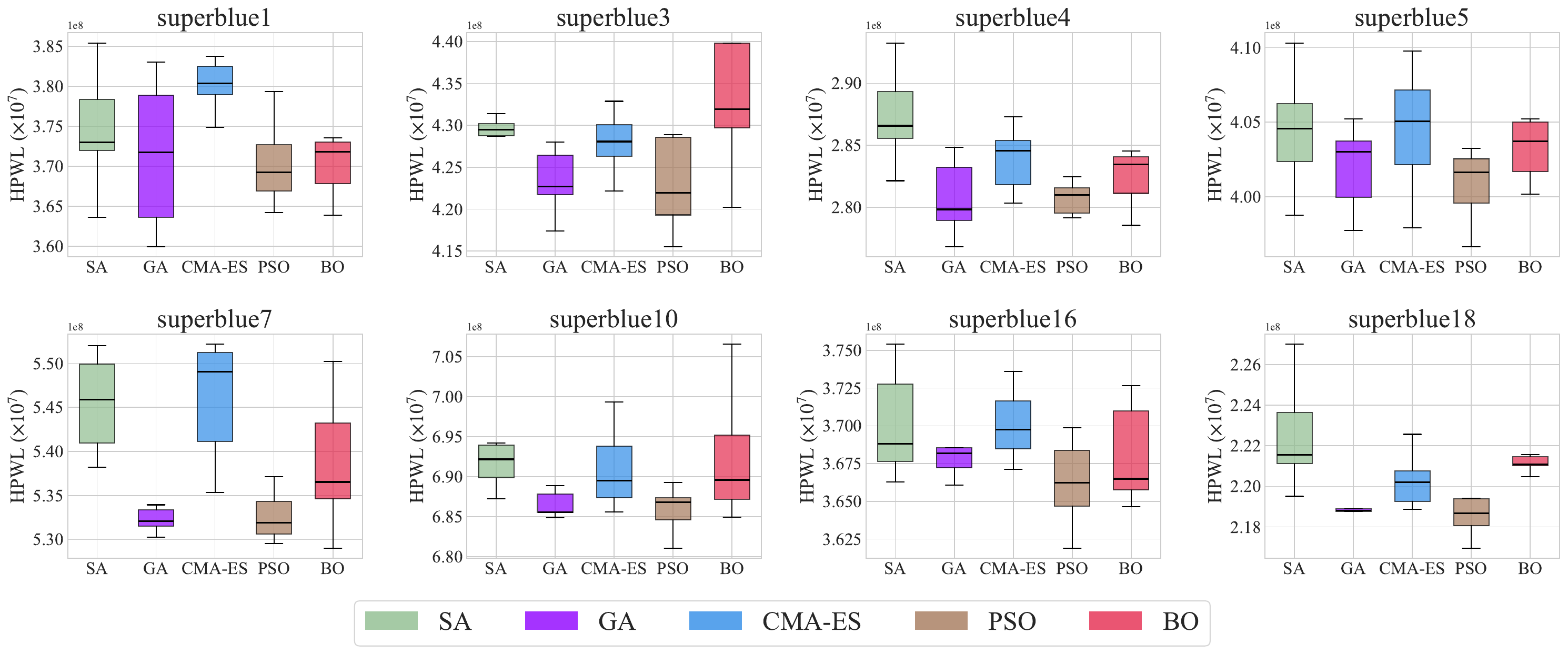}
\caption{Box plots of GP-HPWL produced by HPO methods on eight superblue cases of ICCAD 2015.}
\label{fig:iccad-gp-box}
\end{figure*}

\begin{figure*}[t!]\centering
\includegraphics[width=0.99\linewidth]{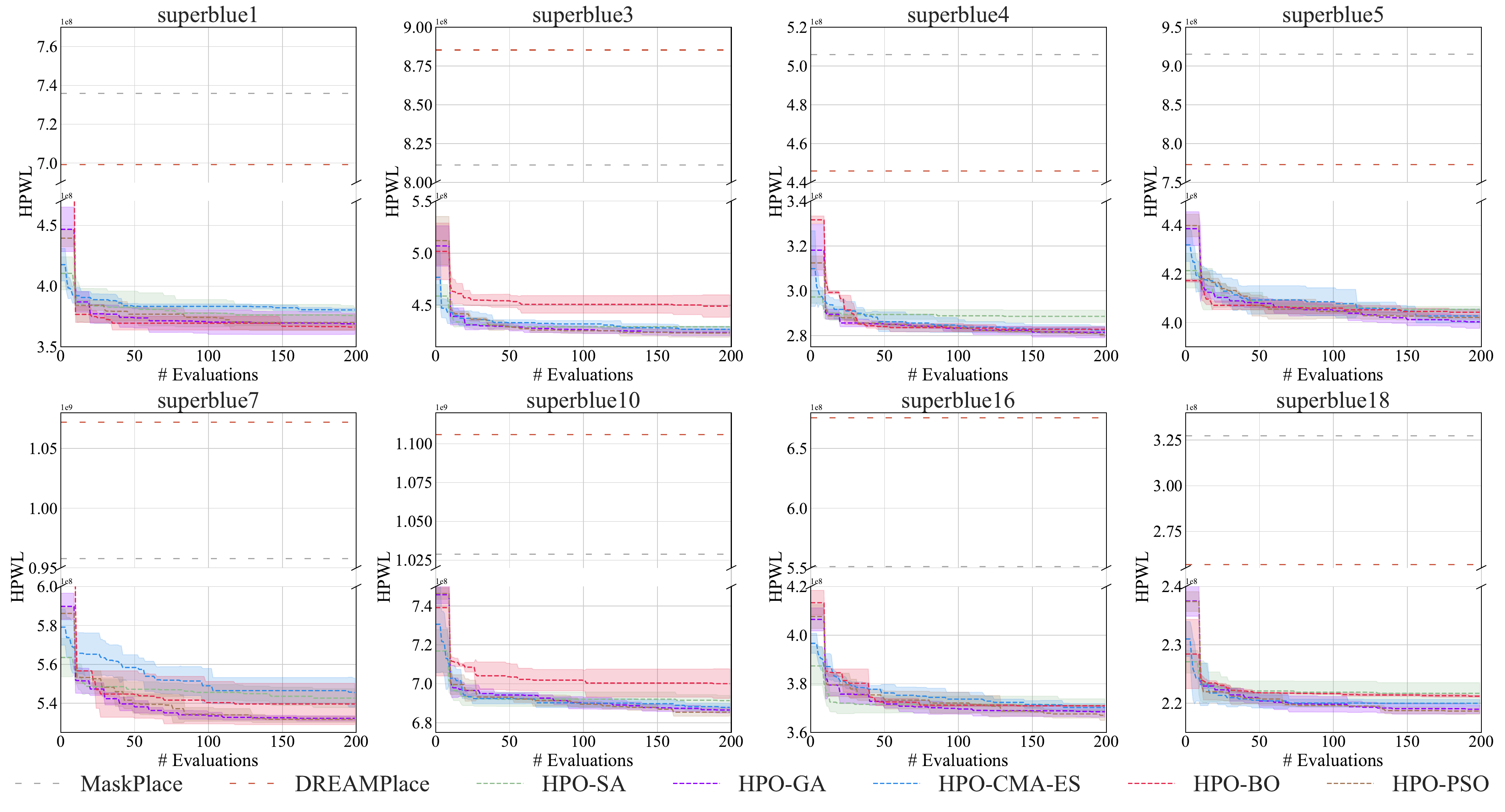}
\caption{GP-HPWL versus the number of evaluations of different methods on ICCAD 2015.}\label{fig:iccad_gp}
\end{figure*}

This subsection provides detailed per-case tables and supplementary curves corresponding to the main-text experiments.
Specifically, we include the detailed MP-HPWL results on ISPD 2005 in Table~\ref{tab:mp-ispd}, the detailed GP-HPWL results on ISPD 2005 in Table~\ref{tab:gp-ispd}, and the detailed MP-HPWL results on ICCAD 2015 in Table~\ref{tab:mp-iccad}.
In addition, we provide the GP-HPWL box plots and the full broken-axis GP-HPWL curves for ICCAD 2015 in Figures~\ref{fig:iccad-gp-box} and~\ref{fig:iccad_gp}.

\begin{table*}[b!]
\caption{MP-HPWL values ($\times 10^5$) obtained by compared methods on the six chip cases of ISPD 2005. Each result consists of the mean and standard deviation of five runs. The results of three RL methods (i.e., AlphaChip~\cite{nature-graph}, MaskPlace~\cite{lai2022maskplace}, EfficientPlace~\cite{fast-place}) are from~\cite{fast-place}. The best and runner-up methods on each chip case are \textbf{bolded} and \underline{underlined}, respectively. The symbols `$\approx$' and `-' indicate that the result is almost equivalent and inferior to the best methods, respectively, according to the Wilcoxon rank-sum test with significance level 0.05.}
\label{tab:mp-ispd}
\resizebox{\textwidth}{!}{
\begin{tabular}{c|c|c|c|c|c|c|c|c}
\toprule
Formulation    & Algorithm & adaptec1 & adaptec2 & adaptec3 & adaptec4 & bigblue1 & bigblue3 & Average Rank  \\
\midrule
\multirow{2}{*}{SP} & SA & 76.80$\pm$3.41 - & 604.18$\pm$12.16 - & 655.61$\pm$20.30 - & 699.85$\pm$1.72 - & 31.73$\pm$0.60 - & 939.84$\pm$32.19 - & 16 \\
                    & GA & 41.80$\pm$3.30 - & 442.77$\pm$13.71 - & 486.91$\pm$10.24 - & 559.43$\pm$6.73 - & 20.04$\pm$0.59 - & 554.93$\pm$23.21 - & 15 \\
\midrule
\multirow{5}{*}{MGO} & SA & 6.32$\pm$0.05 - & 83.61$\pm$5.82 - & 64.05$\pm$0.73 - & 65.53$\pm$0.72 - & 2.44$\pm$0.02 - & 67.51$\pm$3.41 - & 9.67 \\
                     & GA & \textbf{5.80$\pm$0.03 $\approx$} & 61.46$\pm$4.47 - & 56.13$\pm$0.81 - & 56.79$\pm$0.80 - & \underline{2.30$\pm$0.03 -} & 52.40$\pm$2.30 - & 4.83 \\
                     & CMA-ES & 6.28$\pm$0.07 - & 79.43$\pm$6.42 - & 61.69$\pm$0.53 - & 63.28$\pm$1.14 - & 2.42$\pm$0.01 - & 63.84$\pm$4.95 - & 8 \\
                     & PSO & \underline{5.80$\pm$0.02 $\approx$} & 58.78$\pm$4.31 - & 57.64$\pm$0.50 - & 59.56$\pm$0.39 - & 2.35$\pm$0.03 - & 56.04$\pm$3.01 - & 5.83 \\
                     & BO & 6.26$\pm$0.06 - & 80.86$\pm$2.33 - & 62.82$\pm$0.97 - & 62.44$\pm$0.74 - & 2.42$\pm$0.02 - & 67.52$\pm$1.40 - & 8.33 \\
\midrule
\multirow{5}{*}{HPO} & SA & 7.89$\pm$0.12 - & 34.30$\pm$1.82 - & \underline{53.07$\pm$0.89 $\approx$} & 43.33$\pm$0.23 - & 3.45$\pm$0.07 - & 42.44$\pm$1.66 - & 5.33 \\
                     & GA & 7.55$\pm$0.11 - & \underline{32.06$\pm$0.47 $\approx$} & \textbf{52.70$\pm$0.89 $\approx$} & \underline{42.77$\pm$0.54 $\approx$} & 3.35$\pm$0.06 - & \textbf{40.03$\pm$0.72 $\approx$} & 3.5 \\
                     & CMA-ES & 8.15$\pm$0.12 - & 33.00$\pm$0.31 - & 53.57$\pm$0.73 $\approx$ & 43.94$\pm$1.11 - & 3.40$\pm$0.04 - & \underline{41.69$\pm$0.00 -} & 5.67 \\
                     & PSO & 7.99$\pm$0.25 - & \textbf{31.73$\pm$0.70 $\approx$} & 53.07$\pm$0.07 $\approx$ & \textbf{42.49$\pm$0.08 $\approx$} & 3.39$\pm$0.01 - & 42.57$\pm$1.57 - & 4.5 \\
                     & BO & 9.33$\pm$0.76 - & 37.41$\pm$1.82 - & 56.27$\pm$1.59 - & 47.70$\pm$1.22 - & 3.69$\pm$0.04 - & 46.16$\pm$3.51 - & 7.5 \\
\midrule
\multirow{3}{*}{RL} & AlphaChip & 30.01$\pm$2.98 - & 351.71$\pm$38.20 - & 358.18$\pm$13.95 - & 151.42$\pm$13.00 - & 10.58$\pm$1.29 - & 357.48$\pm$47.83 - & 13.83 \\
                    & MaskPlace & 7.62$\pm$0.67 - & 75.16$\pm$4.97 - & 100.24$\pm$13.54 - & 87.99$\pm$3.25 - & 3.04$\pm$0.06 - & 90.04$\pm$4.83 - & 10.33 \\
                    & EfficientPlace & 5.94$\pm$0.04 - & 46.79$\pm$1.60 - & 56.35$\pm$0.99 - & 58.47$\pm$1.61 - & \textbf{2.14$\pm$0.01 $\approx$} & 58.38$\pm$0.54 - & 5.5 \\
\midrule
\multirow{1}{*}{Analytical} & DREAMPlace & 21.20$\pm$9.13 - & 40.05$\pm$4.09 - & 62.94$\pm$0.45 - & 345.37$\pm$39.96 - & 5.40$\pm$0.46 - & 93.02$\pm$6.57 - & 11.67 \\
\bottomrule
\end{tabular}
}
\end{table*}

\begin{table*}[t!]
\caption{GP-HPWL values ($\times 10^7$) obtained by compared methods on the six chip cases of ISPD 2005. Each result consists of the mean and standard deviation of five runs. The results of two RL methods (i.e., MaskPlace~\cite{lai2022maskplace} and EfficientPlace~\cite{fast-place}) are from~\cite{fast-place}. The best and runner-up methods on each chip case are \textbf{bolded} and \underline{underlined}, respectively. The symbols `$\approx$' and `-' indicate that the result is almost equivalent and inferior to the best methods, respectively, according to the Wilcoxon rank-sum test with significance level 0.05.}
\label{tab:gp-ispd}
\resizebox{\textwidth}{!}{
\begin{tabular}{c|c|c|c|c|c|c|c|c}
\toprule 
Formulation    & Algorithm & adaptec1 & adaptec2 & adaptec3 & adaptec4 & bigblue1 & bigblue3 & Average Rank \\
\midrule
\multirow{2}{*}{SP} & SA & 11.87$\pm$0.28 - & 18.29$\pm$0.19 - & 31.51$\pm$0.31 - & 33.50$\pm$0.32 - & 11.54$\pm$0.12 - & 58.34$\pm$0.96 - & 14.83 \\
                    & GA & 11.41$\pm$0.18 - & 17.37$\pm$0.26 - & 30.00$\pm$0.36 - & 33.47$\pm$0.25 - & 11.31$\pm$0.11 - & 51.98$\pm$1.62 - & 13.67 \\
\midrule
\multirow{5}{*}{MGO} & SA & 8.93$\pm$0.09 - & 12.08$\pm$0.50 - & 20.30$\pm$0.47 - & 21.62$\pm$0.26 - & 9.42$\pm$0.06 - & 45.57$\pm$0.56 - & 10.5 \\
                     & GA & 8.49$\pm$0.08 - & 11.05$\pm$0.26 - & 18.45$\pm$0.23 - & 19.80$\pm$0.73 - & 9.29$\pm$0.05 - & 40.43$\pm$0.57 - & 7.83 \\
                     & CMA-ES & 8.98$\pm$0.14 - & 12.25$\pm$0.44 - & 20.00$\pm$0.12 - & 21.93$\pm$0.39 - & 9.45$\pm$0.08 - & 43.27$\pm$0.71 - & 10.67 \\
                     & PSO & 8.65$\pm$0.07 - & 12.27$\pm$0.16 - & 18.01$\pm$0.45 - & 21.35$\pm$0.68 - & 9.32$\pm$0.07 - & 44.52$\pm$1.04 - & 9.33 \\
                     & BO & 8.70$\pm$0.19 - & 12.01$\pm$0.21 - & 17.81$\pm$1.06 - & 20.30$\pm$0.67 - & 9.42$\pm$0.04 - & 45.25$\pm$0.85 - & 9 \\
\midrule
\multirow{5}{*}{HPO} & SA & 6.10$\pm$0.06 $\approx$ & 6.95$\pm$0.12 - & 12.84$\pm$0.10 - & 12.32$\pm$0.09 - & 8.10$\pm$0.05 $\approx$ & 25.36$\pm$0.77 - & 4.67 \\
                     & GA & \textbf{6.05$\pm$0.03 $\approx$} & \underline{6.82$\pm$0.08 $\approx$} & 12.73$\pm$0.11 $\approx$ & \underline{12.12$\pm$0.08 $\approx$} & 8.06$\pm$0.03 $\approx$ & \textbf{24.09$\pm$0.19 $\approx$} & 2 \\
                     & CMA-ES & 6.09$\pm$0.03 $\approx$ & 6.87$\pm$0.14 $\approx$ & \textbf{12.63$\pm$0.08 $\approx$} & 12.21$\pm$0.04 - & 8.11$\pm$0.03 - & 24.72$\pm$0.60 $\approx$ & 3.17 \\
                     & PSO & \underline{6.08$\pm$0.02 $\approx$} & 7.00$\pm$0.12 - & \underline{12.64$\pm$0.08 $\approx$} & 12.18$\pm$0.15 $\approx$ & \underline{8.05$\pm$0.02 $\approx$} & \underline{24.51$\pm$0.40 $\approx$} & 2.67 \\
                     & BO & 6.09$\pm$0.04 $\approx$ & \textbf{6.72$\pm$0.12 $\approx$} & 12.80$\pm$0.12 - & \textbf{12.06$\pm$0.04 $\approx$} & \textbf{8.04$\pm$0.03 $\approx$} & 25.05$\pm$0.33 - & 2.33 \\
\midrule
\multirow{2}{*}{RL} & MaskPlace & 10.86$\pm$0.18 - & 12.98$\pm$0.58 - & 26.14$\pm$0.07 - & 26.14$\pm$0.07 - & 10.64$\pm$0.01 - & 54.98$\pm$1.06 - & 13 \\
                    & EfficientPlace & 7.20$\pm$0.12 - & 9.20$\pm$0.61 - & 16.49$\pm$1.07 - & 14.70$\pm$0.25 - & 8.67$\pm$0.10 - & 28.48$\pm$0.96 - & 6.17 \\
\midrule
\multirow{1}{*}{Analytical} & DREAMPlace & 9.62$\pm$0.78 - & 12.45$\pm$3.31 - & 17.18$\pm$0.51 - & 40.04$\pm$3.87 - & 8.30$\pm$0.06 - & 38.22$\pm$1.48 - & 9.83 \\
\bottomrule
\end{tabular}
}
\end{table*}

\begin{table*}[t!]
\caption{MP-HPWL values ($\times 10^5$) obtained by compared methods on the eight cases of ICCAD 2015. Each result consists of the mean and standard deviation of five runs. The best and runner-up methods on each chip case are \textbf{bolded} and \underline{underlined}, respectively. The symbols `$\approx$' and `-' indicate that the result is almost equivalent and inferior to the best methods, respectively, according to the Wilcoxon rank-sum test with significance level 0.05.}
\label{tab:mp-iccad}
\resizebox{\textwidth}{!}{
\begin{tabular}{c|c|c|c|c|c|c|c|c|c|c}
\toprule
Formulation    & Algorithm & superblue1 & superblue3 & superblue4 & superblue5 & superblue7 & superblue10 & superblue16 & superblue18 & Average Rank \\
\midrule
\multirow{2}{*}{SP} & SA & 12.74$\pm$0.32 - & 30.94$\pm$0.36 - & 20.81$\pm$0.59 - & 55.20$\pm$1.19 - & 24.67$\pm$0.31 - & 11.09$\pm$0.67 - & 29.75$\pm$0.43 - & 6.40$\pm$0.22 - & 14 \\
                    & GA & 5.27$\pm$0.28 - & 13.34$\pm$1.04 - & 11.33$\pm$0.64 - & 31.65$\pm$2.44 - & 14.15$\pm$0.75 - & 2.31$\pm$0.19 - & 14.59$\pm$1.26 - & 2.66$\pm$0.13 - & 12.75 \\
\midrule
\multirow{5}{*}{MGO} & SA & 0.62$\pm$0.01 - & 1.70$\pm$0.03 - & 1.12$\pm$0.02 - & 4.16$\pm$0.07 - & 1.81$\pm$0.03 - & 0.55$\pm$0.00 - & 1.21$\pm$0.04 - & 0.53$\pm$0.01 - & 4.5 \\
                     & GA & \underline{0.59$\pm$0.00 -} & 1.55$\pm$0.01 - & 0.95$\pm$0.01 $\approx$ & \underline{3.84$\pm$0.03 -} & 1.72$\pm$0.02 - & 0.54$\pm$0.00 - & \textbf{0.95$\pm$0.01 $\approx$} & \underline{0.49$\pm$0.00 -} & 2.63 \\
                     & CMA-ES & 0.61$\pm$0.01 - & \underline{1.47$\pm$0.05 -} & 0.97$\pm$0.02 - & 4.53$\pm$0.02 - & \underline{1.37$\pm$0.01 -} & \textbf{0.48$\pm$0.00 $\approx$} & 1.25$\pm$0.01 - & 0.49$\pm$0.01 $\approx$ & 3 \\
                     & PSO & \textbf{0.58$\pm$0.00 $\approx$} & 1.52$\pm$0.00 - & \textbf{0.94$\pm$0.01 $\approx$} & \textbf{3.76$\pm$0.00 $\approx$} & 1.69$\pm$0.02 - & 0.54$\pm$0.00 - & \underline{0.96$\pm$0.01 $\approx$} & \textbf{0.48$\pm$0.00 $\approx$} & 1.88 \\
                     &
                     BO & 0.62$\pm$0.01 - & \textbf{1.39$\pm$0.04 $\approx$} & \underline{0.94$\pm$0.03 $\approx$} & 4.50$\pm$0.10 - & \textbf{1.32$\pm$0.01 $\approx$} & \underline{0.48$\pm$0.00 $\approx$} & 1.19$\pm$0.02 - & 0.50$\pm$0.01 - & 2.38 \\
\midrule
\multirow{5}{*}{HPO} & SA & 2.29$\pm$0.12 - & 4.86$\pm$0.17 - & 2.38$\pm$0.08 - & 10.45$\pm$0.23 - & 3.44$\pm$0.09 - & 1.88$\pm$0.03 - & 3.76$\pm$0.24 - & 1.56$\pm$0.15 - & 9.5 \\
                     & GA & 2.07$\pm$0.14 - & 4.29$\pm$0.19 - & 2.35$\pm$0.10 - & 9.98$\pm$0.13 - & 3.19$\pm$0.03 - & 1.59$\pm$0.05 - & 3.40$\pm$0.12 - & 1.43$\pm$0.09 - & 7.38 \\
                     & CMA-ES & 2.22$\pm$0.09 - & 4.83$\pm$0.24 - & 2.32$\pm$0.11 - & 10.37$\pm$0.38 - & 3.34$\pm$0.07 - & 1.69$\pm$0.10 - & 3.90$\pm$0.31 - & 1.54$\pm$0.15 - & 8.25 \\
                     & PSO & 2.24$\pm$0.05 - & 4.56$\pm$0.42 - & 2.34$\pm$0.09 - & 9.93$\pm$0.07 - & 3.56$\pm$0.05 - & 2.14$\pm$0.18 - & 3.35$\pm$0.03 - & 1.63$\pm$0.19 - & 8.38 \\
                     & BO & 2.60$\pm$0.07 - & 5.93$\pm$0.29 - & 2.66$\pm$0.13 - & 11.77$\pm$0.24 - & 3.93$\pm$0.09 - & 2.74$\pm$0.34 - & 4.51$\pm$0.49 - & 2.01$\pm$0.10 - & 11.13 \\
\midrule
\multirow{1}{*}{RL} & MaskPlace & 2.24$\pm$0.91 - & 4.17$\pm$0.22 - & 1.50$\pm$0.15 - & 9.95$\pm$0.11 - & 2.45$\pm$0.02 - & 0.99$\pm$0.23 - & 2.38$\pm$1.29 - & 1.06$\pm$0.22 - & 6.38 \\
\midrule
\multirow{1}{*}{Analytical} & DREAMPlace & 2.76$\pm$0.06 - & 8.49$\pm$0.36 - & 2.93$\pm$0.03 - & 12.63$\pm$0.39 - & 4.28$\pm$0.04 - & 4.35$\pm$0.06 - & 5.11$\pm$0.10 - & 2.28$\pm$0.02 - & 12.13 \\
\bottomrule
\end{tabular}
}
\end{table*}

\clearpage


\raggedbottom
\makeatletter
\def\@IEEEBIOskipN{0.2\baselineskip}
\makeatother
\newcommand{\rightbiographyskip}{\vspace{0.12in}}

\begin{IEEEbiography}[{\includegraphics[width=0.86in,height=1.08in,clip]{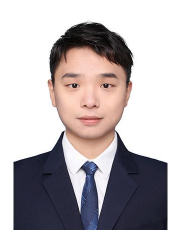}}]{Ke Xue}
received the B.Sc. degree in Mathematics and Applied Mathematics from Sun Yat-Sen University, Guangzhou, China, in 2019, and the Ph.D. degree in Computer Science from Nanjing University, Nanjing, China, in 2026. He is currently a Postdoctoral Researcher with the School of Artificial Intelligence, Nanjing University, and a member of the LAMDA Group. His research interests include machine learning, evolutionary algorithms, and their applications in electronic design automation and AI for Science.
\end{IEEEbiography}

\begin{IEEEbiography}[{\includegraphics[width=0.86in,height=1.08in,clip]{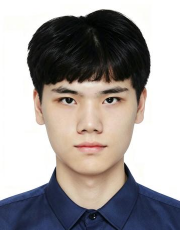}}]{Ruo-Tong Chen}
received the B.Sc. degree in Artificial Intelligence from the School of Artificial Intelligence, Nanjing University, Nanjing, China, in 2025. He is currently pursuing the M.Sc. degree with the School of Artificial Intelligence, Nanjing University, Nanjing, China, supervised by Prof. Chao Qian. His research interests include VLSI physical design, with a focus on floorplanning, placement, timing optimization, and solving these problems with AI techniques, such as reinforcement learning and large language model agent.
\end{IEEEbiography}

\begin{IEEEbiography}[{\includegraphics[width=0.86in,height=1.08in,clip]{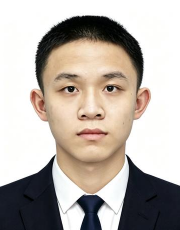}}]{Rong-Xi Tan}
received the B.Sc. degree from the School of Artificial Intelligence, Nanjing University, Nanjing, China, in 2025. He is currently pursuing the Ph.D. degree with the School of Artificial Intelligence, Nanjing University, Nanjing, China, under the supervision of Prof. Chao Qian. He is also a member of the LAMDA Group. His research interests include black-box optimization and large language model agent, and their applications in AI for Science and electronic design automation.
\end{IEEEbiography}

\begin{IEEEbiography}[{\includegraphics[width=0.86in,height=1.08in,clip]{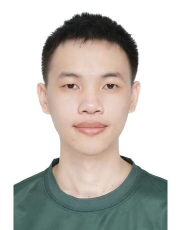}}]{Xi Lin}
received the B.Sc. degree in Artificial Intelligence from Nanjing University, Nanjing, China, in 2025. He is currently pursuing the M.Sc. degree with the School of Artificial Intelligence, Nanjing University. His research interests include VLSI physical design, placement problems, and AI techniques for chip design.
\end{IEEEbiography}

\begin{IEEEbiography}[{\includegraphics[width=0.86in,height=1.08in,clip]{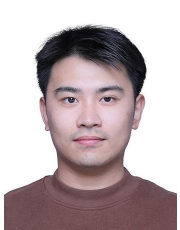}}]{Yunqi Shi}
received the B.Sc. degree in Artificial Intelligence from the School of Artificial Intelligence, Nanjing University, Nanjing, China, in 2023. He is currently pursuing the M.Sc. degree with the School of Artificial Intelligence, Nanjing University, Nanjing, China, supervised by Prof. Chao Qian. His research interests include VLSI physical design, with a focus on placement problems. He won the Best Paper Award at DATE 2025.
\end{IEEEbiography}

\begin{IEEEbiography}[{\includegraphics[width=0.86in,height=1.08in,clip]{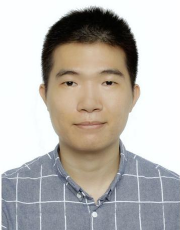}}]{Siyuan Xu}
received his Ph.D. from the University of Texas at Dallas, Richardson, TX, USA, in 2019. Upon graduation, he joined The MathWorks as a Senior Engineer, leading R\&D projects in FPGA-based deep learning systems and HW/SW co-design. He is currently a Senior Researcher at Huawei Noah's Ark Lab, where he focuses on large language models and AI algorithms for the modeling and optimization of next-generation 3D chip physical design, as well as data-driven simulation acceleration.
\end{IEEEbiography}

\rightbiographyskip

\begin{IEEEbiography}[{\includegraphics[width=0.86in,height=1.08in,clip]{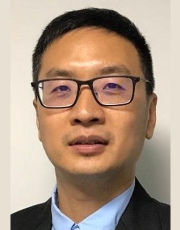}}]{Mingxuan Yuan}
is currently the director and a principal researcher of Huawei Applied AI Research Lab. Before joining Huawei, he worked in HKUST as a post-doc researcher. He obtained his Ph.D degree from the Hong Kong University of Science and Technology, Hong Kong. His research interests include Learning-to-Optimize and Applied AI Models. He has led several research projects including spatio-temporal data analysis, telecommunication data mining, enterprise intelligence, and AI for EDA.
\end{IEEEbiography}

\rightbiographyskip

\begin{IEEEbiography}[{\includegraphics[width=0.86in,height=1.08in,clip]{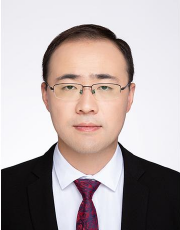}}]{Chao Qian (Senior Member, IEEE)}
received the B.Sc. and Ph.D. degrees in computer science from Nanjing University, Nanjing, China, in 2009 and 2015, respectively. He is currently a Professor in the School of Artificial Intelligence, Nanjing University. His research interests are mainly in artificial intelligence, evolutionary computation, machine learning, AI4Science, and AI4EDA. He has authored the book Evolutionary Learning: Advances in Theories and Algorithms, and has won the ACM GECCO'11 Best Theory Paper Award, the DATE'25 Best Paper Award, and the 21st ACM SIGEVO Humies Bronze Award. He received the 2023 CCF-IEEE CS Young Computer Scientist Award. He serves as a Program Co-Chair of PRICAI'25, and on the editorial board of Artificial Intelligence Journal, Evolutionary Computation Journal, IEEE Trans. Evolutionary Computation, IEEE Computational Intelligence Magazine, etc.
\end{IEEEbiography}

\vfill

\end{document}